\documentclass[final,hidelinks,onefignum,onetabnum]{siamart250211}

\usepackage{lipsum}
\usepackage{amsfonts}
\usepackage{graphicx}
\usepackage{epstopdf}
\usepackage{algorithmic}

\usepackage{float}
\usepackage{xcolor}
\usepackage{amsmath}
\usepackage{amsfonts}
\usepackage{physics}
\usepackage{amssymb}
\usepackage{mathtools}
\usepackage{bm}
\usepackage{multirow}
\usepackage[normalem]{ulem}
\usepackage{url}
\usepackage{bbm}
\usepackage{subcaption}
\usepackage{pifont}

\renewcommand{\H}{\mathcal{H}}

\newcommand{\U}{\mathcal{U}}
\renewcommand{\P}{\mathcal{P}}
\newcommand{\R}{\mathbb{R}}

\newcommand{\V}{\mathcal{V}}
\newcommand{\dint}{\mathrm{d}}
\newcommand{\interior}{\operatorname{int}}

\DeclareMathOperator*{\argmin}{argmin}

\DeclareMathOperator{\TV}{TV}
\DeclareMathOperator{\sinc}{sinc}
\DeclareMathOperator{\mathSpan}{span}

\ifpdf
  \DeclareGraphicsExtensions{.eps,.pdf,.png,.jpg}
\else
  \DeclareGraphicsExtensions{.eps}
\fi

\newsiamremark{remark}{Remark}
\newsiamremark{hypothesis}{Hypothesis}
\crefname{hypothesis}{Hypothesis}{Hypotheses}
\newsiamthm{claim}{Claim}
\newsiamremark{fact}{Fact}
\crefname{fact}{Fact}{Facts}

\headers{Reduced-Order Neural Operator Modeling}{S. Dummer, D. Ye, and C. Brune}

\title{RONOM: Reduced-Order Neural Operator Modeling 
\thanks{\textbf{Funding:} D. Ye and C. Brune acknowledge financial support by NGF AiNed XS Europe under No. NGF.1609.241.020. In addition, S. Dummer and C. Brune acknowledge financial support by the EU EFRO OPoost project under No. OOST-00103.}}

\author{Sven Dummer\thanks{Mathematics of Imaging \& AI, Department of Applied Mathematics, University of Twente, the Netherlands 
  (\email{s.c.dummer@utwente.nl, c.brune@utwente.nl}).}
\and Dongwei Ye\footnotemark[2]
\thanks{Department of Applied Mathematics, School of Mathematics and Physics, Xi'an Jiaotong-Liverpool University, PR China (\email{Dongwei.Ye@xjtlu.edu.cn}).}
\and Christoph Brune\footnotemark[2]
}

\usepackage{amsopn}

\ifpdf
\hypersetup{
  pdftitle={RONOM: Reduced-Order Neural Operator Modeling},
  pdfauthor={S. Dummer, D. Ye, C. Brune}
}
\fi

\usepackage{graphicx}

\begin{document}

\maketitle

\begin{abstract}
Time-dependent partial differential equations are ubiquitous in physics-based modeling, but they remain computationally intensive in many-query scenarios, such as real-time forecasting, optimal control, and uncertainty quantification. Reduced-order modeling (ROM) addresses these challenges by constructing a low-dimensional surrogate model but relies on a fixed discretization, which limits flexibility across varying meshes during evaluation. Operator learning approaches, such as neural operators, offer an alternative by parameterizing mappings between infinite-dimensional function spaces, enabling adaptation to data across different resolutions. Whereas ROM provides rigorous numerical error estimates, neural operator learning largely focuses on discretization convergence and invariance without quantifying the error between the infinite-dimensional and the discretized operators. This work introduces the reduced-order neural operator modeling (RONOM) framework, which bridges concepts from ROM and operator learning. We establish a discretization error bound analogous to those in ROM, and get insights into RONOM's discretization convergence and discretization robustness. Moreover, three numerical examples are presented that compare RONOM to existing neural operators for solving partial differential equations. The results demonstrate that RONOM using standard vector-to-vector neural networks can achieve comparable performance in input generalization and achieves superior performance in both spatial super-resolution and discretization robustness, while also offering novel insights into temporal super-resolution scenarios and ROM-based approaches for learning on time-dependent data.
\end{abstract}



\section{Introduction}
Time-dependent partial differential equations (PDEs) are fundamental for physics-based modeling for many real-world systems \cite{quarteroninumerical}. However, the complexity and scale of these systems often make model evaluations computationally expensive or even intractable. This is especially problematic in many-query scenarios, such as real-time forecasting \cite{NgocCuong2005,biegler2007real,Niroomandi2012}, optimal design and control \cite{biegler2007real,Proctor2016}, and uncertainty quantification \cite{Sudret2017,Ye2022}, which typically require multiple evaluations of the model with varying physical parameters, initial conditions, or boundary conditions. One of the state-of-the-art methods to mitigate this problem is reduced-order modeling (ROM) \cite{quarteroni2015reduced, peherstorfer2016data}. ROM leverages the coherent structure over space, time, or even the parameter space to construct a low-dimensional manifold. The low-dimensional representation of the system resolves the efficiency issue of model evaluation without significantly compromising the accuracy of the model \cite{quarteroni2015reduced}. One common way to construct the manifold is proper orthogonal decomposition (POD) \cite{Rathinam2003}. However, the subspace spanned by the linear bases of POD may suffer from slowly decaying Kolmogorov n-width for transport problems, advection-dominated PDEs, or non-affine parametrizations of domain geometries and physical variables \cite{quarteroni2015reduced}. A variety of nonlinear dimensionality reduction methods have been proposed to address this issue. In particular, deep learning approaches based on autoencoder neural networks have gained considerable attention for their flexibility and strong performance \cite{lee2020model, FRESCA2022114181, farenga2025latent}.

One common limitation of those linear and nonlinear dimensionality reduction methods is that they inherit the spatial discretization from solution snapshots generated by traditional numerical methods, e.g., the finite difference and finite element method (FEM). For instance, the low-dimensional POD representation is formulated in a finite-dimensional subspace spanned by those snapshots, and the bases constructed via low-rank approximation are dependent on the specific discretization of the snapshots \cite{quarteroni2015reduced}. Similarly, the nonlinear manifold in nonlinear model reduction also relies on solutions at a pre-defined discretization. Consequently, these models struggle to generalize across varying discretizations, which is often required in many-query scenarios, e.g., when locally refining grids to resolve fine-scale features. 

One way to mitigate the impact of discretization, but not in the context of ROM, is operator learning \cite{boulle2024mathematical}. Operator learning approximates a mapping between functions. It can be applied to approximate the mapping from an initial condition or source term to the solution of the considered PDE. Such an approximation at the continuous level avoids the explicit dependency on discretization and allows zero-shot super-resolution. Methods such as the random feature model \cite{nelsen2021random}, DeepONet \cite{lu2021learning}, Fourier neural operator (FNO) \cite{li2020fourier}, and graph neural operator (GNO) \cite{li2020neural} have demonstrated their applicability to different PDE problems. Within the broad class of neural operator (NO) methods, several incorporate latent structures. Some use a linear combination of basis functions to map to functions \cite{bhattacharya2021model, hua2023basis, kontolati2023learning}. Others leverage implicit neural representations for more flexible and powerful nonlinear maps \cite{seidman2022nomad, ye2024meta, yin2022continuous, dummer2024rda}.

Whereas neural operators are commonly presented as function-to-function maps, in practice, they work on discretized input functions. The learned operators are usually only discretization-convergent, meaning that the discretized operators converge to a true infinite-dimensional operator as the input resolution is refined. However, as it is only a convergence property, it does not guarantee robustness to discretization changes. The recent ReNO framework \cite{bartolucci2024representation} addresses this limitation by designing operators with a form of equivalence between the input function and its discrete samples. In particular, the CNO \cite{Raonic2023} assumes bandlimited functions and leverages the fact that bandlimited functions satisfy the Nyquist–Shannon theorem, which ensures equivalence between functions and their point samples. Consequently, any input over the Nyquist rate can be projected back to the same rate of training before being passed to the operator, yielding equal performance when tested on higher-resolution scenarios. But on the other hand, such a setting also limits its applicability. 

While the properties of neural operators, such as discretization convergence and discretization invariance, are widely explored in existing literature, it often lacks a thorough analysis of how well the discretized operators approximate their underlying infinite-dimensional counterparts. 
To the best of our knowledge, only Lanthaler et al.~\cite{lanthaler2024discretization} investigate this error by analyzing the discretization errors made by FNOs. 
In contrast, several ROM methods are equipped with rigorous error analysis, guaranteeing the accuracy of the reduced system's numerical construction and approximation \cite{quarteroni2015reduced}. This is closely related to discretization convergence and discretization invariance. If the error estimates guarantee that refining the numerical solution brings it closer to the true solution, then the numerical solver is discretization convergent. 

\begin{figure}[t]
    \centering
    \includegraphics[width=1\textwidth]{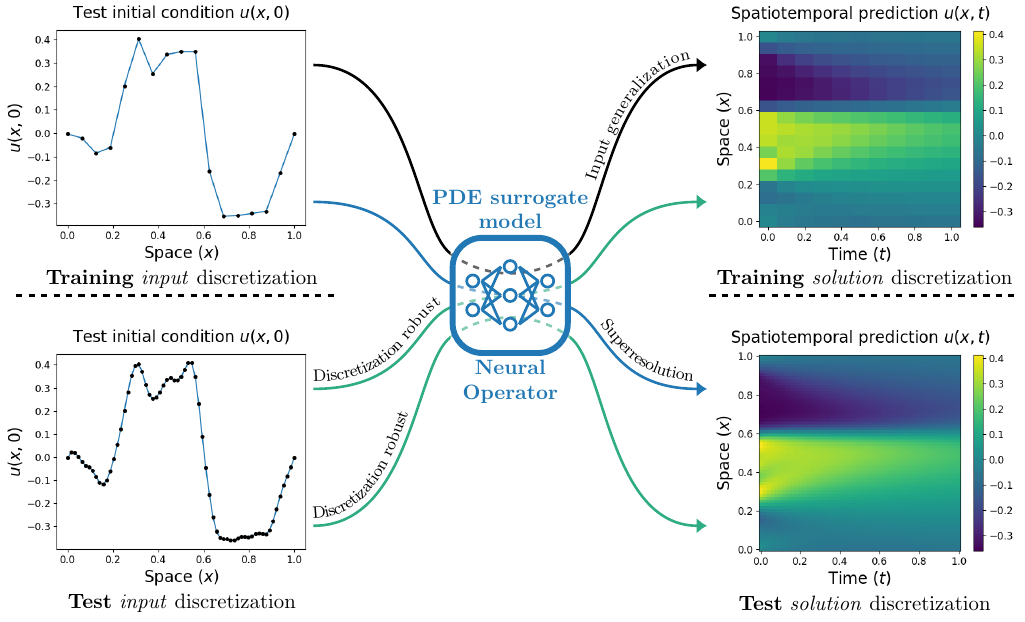}
\caption{\textbf{\textit{Robustness and superresolution through RONOM}}. Traditional finite-dimensional ROM methods for PDE surrogate modeling can generalize to new inputs, provided these input functions use the same fixed discretization as in training (black arrow). In contrast, RONOM can also enhance the resolution beyond the training resolution (blue arrow), and it remains applicable and robust when inputs are given at different discretizations than those used in training (green arrows).}
\label{fig:different_tasks}
\end{figure}

This work proposes the reduced-order neural operator modeling (RONOM) framework that connects operator learning and ROM. The general structure of RONOM is illustrated in Figure~\ref{fig:overview_RONOM}. An architecture based on standard vector-to-vector neural networks is adopted, inspired by the kernel method of Batlle et al.~\cite{batlle2024kernel}. We demonstrate that appropriate lifting and sampling turn these standard neural network structures, which also appear in ROM, into a neural operator that can achieve competitive performance.
We prove a discretization error bound for RONOM and evaluate its performance in terms of the three desired properties illustrated in Figure \ref{fig:different_tasks}, namely generalization to unseen inputs, super-resolution, and discretization robustness. Moreover, it is important to note that the latent code in this work is not necessarily a reduced representation in the traditional ROM sense. Since functions are considered and their discrete resolution may vary, it is more accurate to view the latent code as a finite-dimensional representation (projection) of the function at arbitrary resolution. The main contributions of this work are summarized as follows:

\begin{figure}[t]
    \centering
    \includegraphics[width=0.85\linewidth]{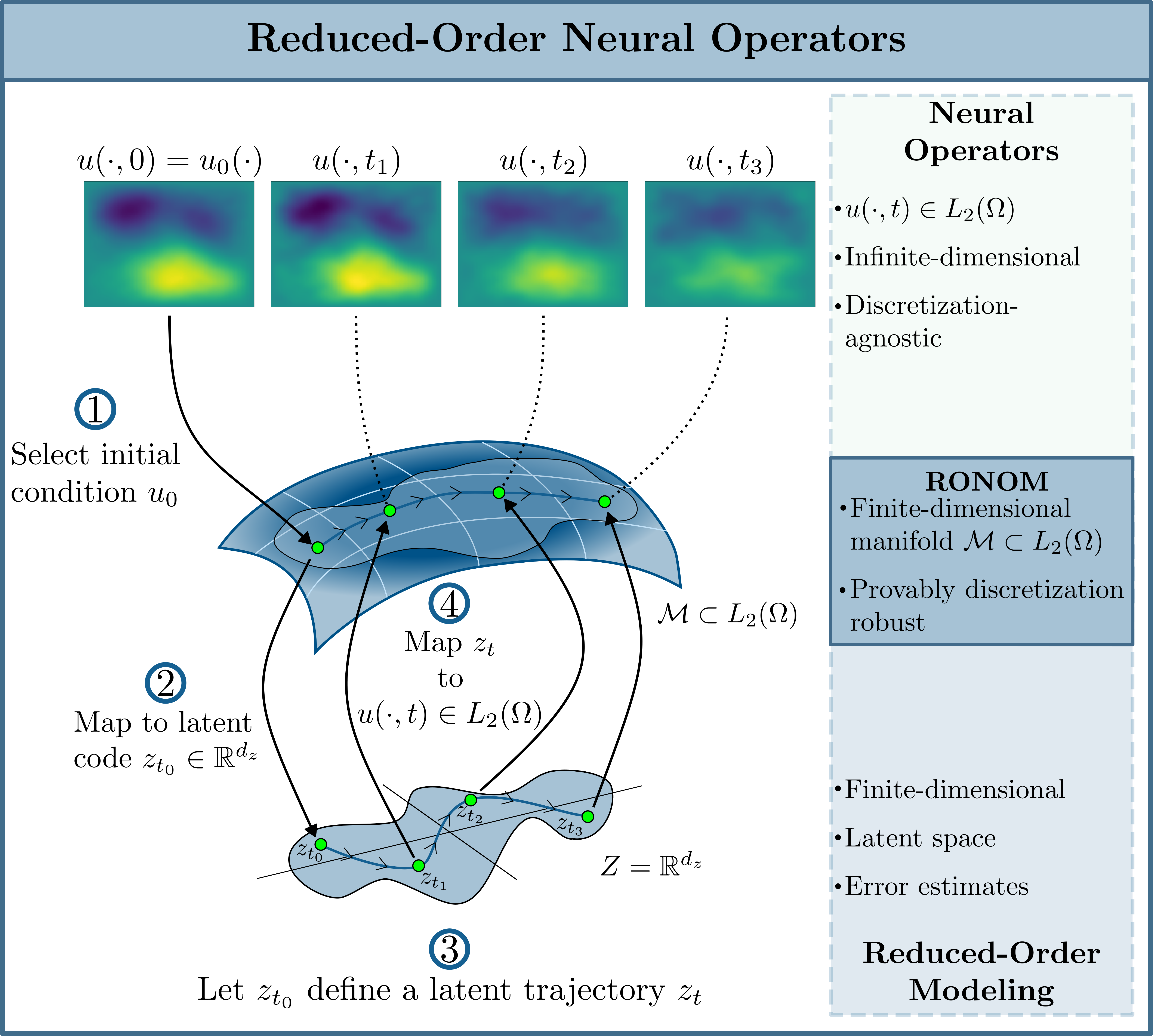}
    \caption{\textbf{\textit{RONOM}} is a modeling framework that combines neural operators and reduced order modeling. It first maps the initial condition into a latent space. From this initial latent representation, a latent trajectory is obtained over time. The full trajectory of functions is recovered by decoding the latent codes at each time instance.}
    \label{fig:overview_RONOM}
\end{figure}

\begin{itemize}
    \item \textbf{Modeling}: a discretization-convergent neural operator is proposed that connects traditional ROM to recent developments in neural operator learning. 
    \item \textbf{Analysis}: discretization error bounds are established for the neural operator, unifying numerical error analysis in ROMs with the discretization convergence concept of neural operators. 
    \item \textbf{Numerics}: numerical experiments provide insight into the temporal super-resolution of operator learning and into ROM-based approaches for learning on time-dependent data. Moreover, the experiments demonstrate that \textit{i)} comparable input generalization performance can be achieved compared to FNO, CNO, and DeepONet, and \textit{ii)} spatial super-resolution and discretization robustness are achieved in cases where FNOs and CNOs fail. 
\end{itemize}

\section{The reduced-order neural operator model}
This section introduces the details of RONOM. The specific time-dependent PDEs that RONOM is designed to solve are first described. The model architecture, including the encoder, decoder, and neural ordinary differential equation (ODE) components, is subsequently detailed. Discretization error estimates and Lipschitz continuity results are also presented. These results are crucial for providing RONOM's error estimates in Section \ref{sec:discretization_error_estimate}. 

\subsection{Problem statement}
Consider a time-dependent PDE on a bounded domain $\Omega \subset \mathbb{R}^d$,
\begin{equation}\label{eq:pde}
\begin{aligned}
 \frac{\partial}{\partial t}u(\bm{x},t) + \mathcal{N}(u,\bm{x}) = 0,& \hspace{3.5cm} (\bm{x},t) \in \Omega \times [0,T]   \\
\mathcal{B} u(\bm{x},t) = g(\bm{x}),& \hspace{3.5cm} (\bm{x},t) \in \partial \Omega \times [0,T]\\
u(\bm{x},0)  = u_0(\bm{x}),& \hspace{3.5cm} \bm{x} \in \Omega.
\end{aligned}
\end{equation}
where $u$ denotes the solution of the PDE; $\mathcal{N}$ is a general differential operator and $\mathcal{B}$ is the boundary condition operator defined on the domain boundary $\partial \Omega$. In this work, we aim to learn the nonlinear solution operator $\mathcal{K}\colon (u_0, t) \mapsto u(\cdot, t)$ based on supervised training data. 

\subsection{Overview of RONOM}
RONOM aims to approximate the solution operator $\mathcal{K}$ by an operator $\mathcal{K}^{\dagger}$ consisting of three maps,
\begin{equation*}
    \mathcal{K}(u_0, t) \approx \mathcal{K}^{\dagger}(u_0, t) \coloneqq (D\circ \mathcal{F}_t \circ E)(u_0).
\end{equation*}
The map $E\colon u_0 \mapsto \mathbf{z}_0 \in \mathbb{R}^{d_z}$ is the encoder that maps the initial condition to a finite-dimensional latent code of dimension $d_z$. Such a latent space allows us to explore and march over time, which is analogous to the reduced system of the reduced basis method \cite{quarteroni2015reduced, NgocCuong2005}. Our model realizes the time marching in latent space through the flow operator $\mathcal{F}_t: \mathbb{R}^{d_z} \to \mathbb{R}^{d_z}$. More specifically, $\mathcal{F}_t(\mathbf{z}_0) = \mathbf{z}(t)$ with $\mathbf{z}(t)$ the solution of a first-order ODE system,
\begin{equation}\label{eq:latent_ode}
    \frac{\mathrm{d}\mathbf{z}(t)}{\mathrm{d}t} = \mathbf{v}(\mathbf{z}(t),t),\quad \mathbf{z}(0) = \mathbf{z}_0 =E(u_0),
\end{equation}
where $\mathbf{v}:\mathbb{R}^{d_z} \to \mathbb{R}^{d_z}$ determines the latent dynamics. Finally, the decoder $D: \mathbb{R}^{d_z} \to L_2(\Omega)$ recovers the approximated solution from the latent code $\mathbf{z}(t)$. 

Each component of $\mathcal{K}^{\dagger}$ is characterized by neural networks. Given the available training data $\{u_{i0},\{u_{ij}\coloneq u_i(t_j)\}_{j=0}^{N_t-1} \}_{i=1}^{N_{s}}$, the learning of the solution operator $\mathcal{K}^{\dagger}$ relies on minimizing the objective function,
\begin{equation*}
\begin{aligned}
    \min_{\bm{\theta}} \quad &  \frac{1}{N_s N_t}\sum_{i=1}^{N_s} \sum_{j=0}^{N_t - 1} 
    \norm{u_{ij} - \widehat{u}_{i}(t_j)}_{L_2(\Omega)}^2 \\
    \textrm{s.t. } \quad \, \, \,  & \frac{\dint}{\dint t} \bm{\mathrm{z}}_i(t) = \mathbf{v}(\bm{\mathrm{z}}_i(t), t), \quad \bm{\mathrm{z}}_i(0) = E(u_{i0}), \quad \widehat{u}_i(t)=D(\bm{\mathrm{z}}_i(t)), \\
\end{aligned}
\end{equation*}


\subsection{Encoder}
To ensure the discretization convergence and discretization robustness of the encoder, inspiration is drawn from the optimal recovery approach presented in Batlle et al. \cite{batlle2024kernel} and the frame sequence approach employed in the CNO \cite{Raonic2023}. Both approaches map the input samples to a full function, followed by the application of a predefined linear operator to obtain a finite-dimensional vector representation. 

Our encoder first uses a regularized $L_2$ projection $\mathcal{P}_\mathcal{V}^\lambda$ to project the input function to a finite-dimensional subspace $\mathcal{V}:=\mathSpan\{\phi_1, \phi_2, \cdots, \phi_{N_b} \} \subset L_2(\Omega)$, where $\{\phi_i\}_{i=1}^{N_b}$ are linearly independent. In particular, the projection $\mathcal{P}_\mathcal{V}^\lambda$ is written as,
\begin{equation}
    \mathcal{P}_\mathcal{V}^\lambda f := \argmin_{u \in \mathcal{V}} \int_\Omega |u(\bm{x}) - f(\bm{x})|^2 \dint \bm{x} + \lambda \norm{u}_\V^2,
    \label{eq:reg_proj}
\end{equation}
with $\norm{\cdot}_\V$ a norm on $\V$ coming from an inner product $\braket{\cdot}{\cdot}_\V$. To compute this projection for discretized input functions in the form of grids, meshes, or point clouds, a discretize-then-optimize approach is applied, which is discussed in Section \ref{sec:disc_then_opt_proj}. After the projection, we apply a measurement operator $M\colon \mathcal{V} \to \mathbb{R}^{d_m}$, for instance by sampling the projection at specific points or by mapping it to the basis coefficients \cite{bhattacharya2021model, hua2023basis, bahmani2025resolution}. Subsequently, the resulting vector is processed with a neural network $\mathcal{E}_\varphi \colon \mathbb{R}^{d_m} \to \mathbb{R}^{d_z}$. Summarizing, the full (infinite-dimensional) encoder is given by:
\begin{equation}
    E(u_0) := (\mathcal{E}_\varphi \circ M\circ\mathcal{P}_\mathcal{V}^\lambda)(u_0).
    \label{eq:encoder_formula}
\end{equation}
For a discussion on how this encoder connects to the frame-sequence approach used by the CNO, see \ref{app:connection_encoder_and_CNO}. Note that it is important to ensure that the measurement operator $M$ is injective when restricted to the subspace $\mathcal{V}$. This injectivity guarantees the equivalence between the measurement values and the (projected) functions in $\V$. To ensure injectivity, the matrix $\mathrm{M}_\phi \in \R^{d_m \times N_b}$ with entries $(\mathrm{M}_\phi)_{\cdot i} = M\phi_i$, must have full rank $N_b$, which implies the condition $d_m \geq N_b$. Therefore, the basis functions spanning $\mathcal{V}$ should be chosen such that $N_b \leq d_m$, and the measurement operator $M$ must be designed such that the matrix $\mathrm{M}_\phi$ has rank $N_b$.

\begin{remark}
    Our encoder is closely related to encode-process-decode FNO approaches for complex geometries \cite{li2023fourier, li2023geometry}. Our encoder first reconstructs an approximation of the underlying function from samples and then applies a measurement operator to produce a consistent vector representation, such as function values on a latent grid. In contrast, the two approaches by Li et al.~\cite{li2023fourier, li2023geometry} directly map sampled values to a latent grid by either deforming the domain or mesh with a (learned) deformation, or by applying a convolution operator discretized over local neighborhoods. A similar distinction exists between the decoding in the two approaches by Li et al.~\cite{li2023fourier, li2023geometry} and our decoding process, which is described in Section \ref{sec:decoder}.
\end{remark}

\subsubsection{Discretized projection} \label{sec:disc_then_opt_proj}
To apply the encoder in Equation \eqref{eq:encoder_formula} for data in the form of a mesh or point cloud, the $\mathcal{P}_\mathcal{V}^\lambda$ mapping is discretized using a discretize-then-optimize approach. The discretized projection solves a regularized least-squares problem that maps the input $\{\bm{x}_i, f(\bm{x}_i)\}_{i=1}^n = \{\bm{x}_i, f_i\}_{i=1}^n$ to a function
\begin{equation}
    \widehat{\mathcal{P}}_\V^\lambda(\{\bm{x}_i, f_i\}_{i=1}^n) \coloneqq \argmin_{u \in \V} \sum_{i=1}^n w_i|u(\bm{x}_i) - f_i|^2 + \lambda \norm{u}_\V^2, 
    \label{eq:discr_proj}
\end{equation}
where $w_i\geq0$ are weighting constants used to approximate the data fidelity integral in Equation \eqref{eq:reg_proj}. The convergence of the solution of Equation \eqref{eq:discr_proj} to the solution of Equation \eqref{eq:reg_proj} will be discussed in the next section. The following theorem shows that the regularized projection can be interpreted as a specific regularized instance of the kernel approach in Batlle et al.\ \cite{batlle2024kernel}.
\begin{theorem}
      Let $\tilde{\phi}_i$ be an orthonormal basis of the Hilbert Space $(\mathcal{V}, \braket{\cdot}{\cdot}_{\V})$ with $\V\coloneqq \mathSpan\{\phi_1, \phi_2, \cdots, \phi_{N_b} \}$. Define  $K\colon \Omega \times \Omega \to \R$ as $K(\bm{x},\bm{y}) = \sum_{i=1}^{N_b} \tilde{\phi}_i(\bm{x}) \tilde{\phi_i}(\bm{y})$. Then $\left(\V, \braket{\cdot}_\V\right)$ is a reproducing kernel Hilbert space with kernel $K$ and optimization problem \eqref{eq:discr_proj} the soft-constraint version of the optimal recovery problem.
      \label{thm:vv_rkhs_structure_mathcal_V}
\end{theorem}
\begin{proof}
    To show $K$ is the reproducing kernel of $\V$, note that
    \begin{align*}
            \braket{K(\cdot, \bm{x})}{\tilde{\phi}_j}_{\V}  = 
            \sum_{i=1}^{N_b} \tilde{\phi}_i(\bm{x}) \braket{\tilde{\phi_i}}{\tilde{\phi}_j}_{\V}= 
            \sum_{i=1}^{N_b} \tilde{\phi}_i(\bm{x}) \delta_{ij} = \tilde{\phi}_j(\bm{x}).
    \end{align*}
    Therefore for any $f = \sum_{i=1}^{N_b}a_i\tilde{\phi}_i \in \V$, $a_i\in \R$, the following holds,
    \begin{align*}
            \braket{K(\cdot, \bm{x})}{f}_{\V} =
            \sum_{i=1}^{N_b}a_i \braket{K(\cdot, \bm{x})}{\tilde{\phi}_i}_{\V} = 
            \sum_{i=1}^{N_b} a_i \tilde{\phi}_i(\bm{x}) = 
            f(\bm{x}),
    \end{align*}
    which confirms that $K$ serves as the reproducing kernel of $\mathcal{V}$ with respect to the inner product $\braket{\cdot}{\cdot}_{\V}$. 
\end{proof}
\begin{remark}
    Thanks to the RKHS structure in Theorem \ref{thm:vv_rkhs_structure_mathcal_V}, the solution admits two possible expansions:
\begin{equation*}
    u(x) = \sum_{i=1}^n K(x, x_i)\, a_i, \quad \text{or} \quad u(x) = \sum_{i=1}^{N_b} a_i \phi_i(x),
\end{equation*}
where the first expansion follows from the RKHS representer theorem, and the second is by definition of $\mathcal{V}$.

When $n < N_b$, the kernel expansion can be more efficient, as it leads to a system of size $n$ rather than $N_b$. However, it requires computing an orthonormal basis $\{\tilde{\phi}_i\}_{i=1}^{N_b}$ and evaluating the kernel using this basis, both of which can be computationally expensive. In such cases, one can instead use the $\phi_i$'s and construct a system of size $N_b$. Conversely, when $N_b < n$, the $\phi_i$-basis expansion is also preferable, as it reduces the number of terms. Motivated by this discussion and the simplicity of the method, we adopt the $\phi_i$-basis expansion in the next section to (approximately) solve the discrete regularized projection in Equation \eqref{eq:discr_proj}.
\end{remark}

\subsubsection{Error estimate projection's basis coefficients}
To get an encoder error estimate in Section \ref{sec:discretization_error_estimate}, it is necessary to bound the difference between the solutions to \eqref{eq:reg_proj} and \eqref{eq:discr_proj}. 
By expressing $u(\bm{x}) = \sum^{N_b}_{i=1} \alpha_i \phi_i(\bm{x}) = \bm{\phi}(\bm{x})^{\top}\bm{\alpha}$, the problems \eqref{eq:reg_proj} and \eqref{eq:discr_proj} can be reformulated as follows,
\begin{equation}
    \min_{\bm{\alpha} \in \R^{N_b}} \bm{\alpha}^{\top} \Phi \bm{\alpha} - 2 \bm{\alpha}^{\top} \braket{f}{\bm{\phi}}_{L_2(\Omega)} + \norm{f}_{L_2(\Omega)}^2 + \lambda \bm{\alpha}^{\top} \mathrm{L} \bm{\alpha},
    \label{eq:opt_prob_cond_proj}
\end{equation}
and
\begin{equation}
    \min_{\bm{\alpha} \in \R^{N_b}} \bm{\alpha}^{\top} \Tilde{\Phi} \mathrm{D}_w \Tilde{\Phi}^{\top} \bm{\alpha} - 2 \bm{\alpha}^{\top} \Tilde{\Phi} \mathrm{D}_w \bm{f} + \bm{f}^{\top} \mathrm{D}_w \bm{f} + \lambda \bm{\alpha}^{\top} \mathrm{L} \bm{\alpha},
    \label{eq:opt_prob_cond_discr_proj}
\end{equation}
respectively, where $(\mathrm{L})_{ij}=\braket{\phi_i}{\phi_j}_\V$, $\Tilde{\Phi}_{ij} = \phi_i(\bm{x}_j)$, $(\braket{f}{\bm{\phi}}_{L_2(\Omega)})_i = \braket{f}{\phi_i}_{L_2(\Omega)}$, $(\mathrm{D}_w)_{ij} = w_i \delta_{ij}$, and $\bm{f} = [f_1,\cdots,f_n]^{\top}$. Hence, the difference between the solutions is determined by the difference between the basis coefficients. As shown in the next theorem, this is mainly controlled by how well $\Tilde{\Phi} \mathrm{D}_w \Tilde{\Phi}^\top$ approximates $\Phi$ and how well $\Tilde{\Phi}\mathrm{D}_w \bm{f}$ approximates $\braket{f}{\bm{\phi}}_{L_2(\Omega)}$,
\begin{theorem}
    Assume $\bm{\alpha}^\dagger$ solves Equation \eqref{eq:opt_prob_cond_proj} and $\bm{\alpha}$ Equation \eqref{eq:opt_prob_cond_discr_proj}. Assuming $\mathrm{L}$ is invertible, the difference between them can be bounded by:
    \begin{equation*}
    \begin{aligned}
        \norm{\bm{\alpha} - \bm{\alpha}^\dagger}_{2} &\leq \frac{ \norm{\mathrm{L}^{-1}}_2}{\lambda}\Big(\norm{\braket{f}{\bm{\phi}}_{L_2(\Omega)} - \Tilde{\Phi} \mathrm{D}_w \bm{f}}_2 \\ &+ \norm{(\Tilde{\Phi} \mathrm{D}_w \Tilde{\Phi}^{\top} - \Phi)}_2 \norm{(\Phi + \lambda \mathrm{L})^{-1}}_2 \norm{\braket{f}{\bm{\phi}}_{L_2(\Omega)}}_2 \Big).
    \end{aligned}
    \end{equation*}
    \label{lemma:general_coeff_error_estimate}
\end{theorem}
\begin{proof}
 Putting the gradient with respect to $\bm{\alpha}$ to zero in problems \eqref{eq:opt_prob_cond_proj} and \eqref{eq:opt_prob_cond_discr_proj} gives us $(\Phi + \lambda \mathrm{L}) \bm{\alpha}^\dagger = \braket{f}{\bm{\phi}}_{L_2(\Omega)}$ and $(\Tilde{\Phi} \mathrm{D}_w \Tilde{\Phi}^{\top} + \lambda \mathrm{L}) \bm{\alpha} = \Tilde{\Phi} \mathrm{D}_w \bm{f}$. Moreover,
    \begin{equation*}
        (\Tilde{\Phi} \mathrm{D}_w \Tilde{\Phi}^{\top} + \lambda \mathrm{L})(\bm{\alpha}^\dagger - \bm{\alpha})  = (\Phi + \lambda \mathrm{L})(\bm{\alpha}^\dagger) - (\Tilde{\Phi} \mathrm{D}_w \Tilde{\Phi}^{\top} + \lambda \mathrm{L}) \bm{\alpha} + (\Tilde{\Phi} \mathrm{D}_w \Tilde{\Phi}^{\top} - \Phi) \bm{\alpha}^\dagger
    \end{equation*}
    By the optimality conditions above, this becomes
    \begin{equation}
        \left(\braket{f}{\bm{\phi}}_{L_2(\Omega)} - \Tilde{\Phi} \mathrm{D}_w \bm{f}\right) + (\Tilde{\Phi} \mathrm{D}_w \Tilde{\Phi}^{\top} - \Phi) (\Phi + \lambda \mathrm{L})^{-1} \braket{f}{\bm{\phi}}_{L_2(\Omega)}.
        \label{eq:lemma_expansion}
    \end{equation}
    To obtain a bound on the difference $\bm{\alpha} - \bm{\alpha}^\dagger$, note that
    \begin{equation*}
        \begin{split}
            \norm{\bm{\alpha} - \bm{\alpha}^\dagger}_2 & = \norm{ (\Tilde{\Phi} \mathrm{D}_w \Tilde{\Phi}^{\top} + \lambda \mathrm{L})^{-1}  (\Tilde{\Phi} \mathrm{D}_w \Tilde{\Phi}^{\top} + \lambda \mathrm{L}) (\bm{\alpha} - \bm{\alpha}^\dagger) }_2 \\
            & \leq  \norm{(\Tilde{\Phi} \mathrm{D}_w \Tilde{\Phi}^{\top} + \lambda \mathrm{L})^{-1}}_2 \norm{ (\Tilde{\Phi} \mathrm{D}_w \Tilde{\Phi}^{\top} + \lambda \mathrm{L}) (\bm{\alpha} - \bm{\alpha}^\dagger) }_2 .
        \end{split}
    \end{equation*}
    This shows that $\mathrm{L}$ is positive definite and therefore invertible. Since $\Tilde{\Phi} \mathrm{D}_w \Tilde{\Phi}^{\top}$ is positive semi-definite and $\mathrm{L}$ is positive definite, we find that \cite[Theorem 4.2.6]{horn2012matrix} yields
    \[
        \gamma_{\text{min}}\left(\Tilde{\Phi} \mathrm{D}_w \Tilde{\Phi}^{\top} + \lambda \mathrm{L} \right) = \min_{\bm{\alpha} } \bm{\alpha}^\top \left( \Tilde{\Phi} \mathrm{D}_w \Tilde{\Phi}^{\top} + \lambda \mathrm{L} \right) \bm{\alpha}  \geq \lambda \min_{\bm{\alpha}}\bm{\alpha}^\top \mathrm{L} \bm{\alpha} = \lambda \gamma_{\text{min}} \left(\mathrm{L}\right)
    \]
    with $\gamma_{\text{min}}$ denoting the smallest eigenvalue and $\lambda > 0$. Hence, for any $\lambda>0$,
    \[
        \left\lVert\left(\Tilde{\Phi} \mathrm{D}_w \Tilde{\Phi}^{\top} + \lambda \mathrm{L}\right)^{-1}\right \rVert_2 = \frac{1}{ \gamma_{\text{min}}\left(\Tilde{\Phi} \mathrm{D}_w \Tilde{\Phi}^{\top} + \lambda \mathrm{L} \right)} \leq \frac{1}{\lambda}\frac{1}{\gamma_{\text{min}}\left(\mathrm{L}\right)} = \frac{\lVert \mathrm{L}^{-1} \rVert_2}{\lambda}.
    \]
    Using this estimate and the triangle inequality used on \eqref{eq:lemma_expansion} completes the proof.
\end{proof}
As stated in the theorem, it suffices to quantify how accurately $\Tilde{\Phi} \mathrm{D}_w \Tilde{\Phi}^T$ approximates $\Phi$, as well as how well $\Tilde{\Phi}\mathrm{D}_w \bm{f}$ approximates $\braket{f}{\bm{\phi}}_{L_2(\Omega)}$. To this end, note that
\begin{equation*}
    \begin{aligned}
    \left( \Tilde{\Phi} D_w \Tilde{\Phi}^{\top} \right)_{ij} = \sum_{k=1}^{N_b} w_k \phi_i(\bm{x}_k) \phi_j(\bm{x}_k), \quad \left( \Tilde{\Phi} \mathrm{D}_w \bm{f} \right)_{i} = \sum_{k=1}^{N_b} w_k \phi_i(\bm{x}_k) f(\bm{x}_k).
    \end{aligned}
\end{equation*}
Therefore, it suffices to establish error estimates for $L_2$ inner products. 

\subsubsection{Integral discretization error estimates}\label{sec:error_estimates_integral_discretization}
Given a mesh, the standard approach to approximate the (inner product) integrals is numerical integration, also known as numerical quadrature. To understand and assess the impact of quadrature errors on the approximation of the $L_2$ inner products, an error bound is presented below. The detailed proof is provided in Appendix \ref{app:integral_error_estimate}.

\begin{theorem}
    Assume that the compact domain $\Omega$ is approximated by a mesh $\Omega^N \subseteq \Omega$ consisting of $N$ mesh elements $\Omega_k$ with nonempty interior, i.e., $\interior(\Omega_k)$: $\Omega^N = \bigcup_{k=1}^N \Omega_k$ and $\interior(\Omega_k) \cap \interior(\Omega_{\tilde{k}}) = \emptyset$ for $k\neq \tilde{k}$. Assume either:
    \begin{itemize}
        \item $f \in C^{p+1}(\Omega)$ and $\{w_{i,k}, \bm{x}_{i,k}\}_{i=1}^{m_k}$ is a quadrature rule of order $p$, meaning $\int_{\Omega_k} q(\bm{x}) \dint \bm{x} = \sum_{i=1}^m w_{i,k} q(\bm{x}_{i,k})$ for all $p$-th order polynomials $q$ on $\Omega_k$.
        \item $f \in \TV(\Omega)$, $w_{i,k} = |\Omega_k| \widetilde{w}_{i,k}$ with $\sum_{i=1}^{m_k}\widetilde{w}_{i,k} = 1$, and $\bm{x}_{i,k} \in \Omega_k$.
    \end{itemize}
    Then $\left|\int_{\Omega} f(\bm{x}) \dint \bm{x} -  \sum_{k=1}^N \left( \sum_{i=1}^{m_k} w_{i,k} f(\bm{x}_{i,k}) \right) \right|$ is bounded by
    \begin{equation*} 
 |\Omega \Delta \Omega^N| \norm{f}_{L_\infty(\Omega)} + \begin{cases}
                c h^{p+1} |\Omega^N| \sup_{\substack{\bm{x} \in \Omega \\ \sum_{i=1}^d \gamma_i = p+1}} \left|\partial^\gamma f(\bm{x}) \right|, & \quad f \in C^{p+1}(\Omega) \\
                h \TV(f), & \quad f \in \TV(\Omega),
        \end{cases}
    \end{equation*}
    where $c \in \R$ only depends on $p$, $\partial^\gamma f(\bm{x}) := 
    \frac{\partial^{|\gamma|}}{\partial x_1^{\gamma_1} \cdots \partial x_d^{\gamma_d}} f(\bm{x})$ with a multi-index notation $\gamma=(\gamma_1, \gamma_2, \cdots, \gamma_d)$, $h_{\Omega_k}\coloneqq \sup_{\bm{x}_1,\bm{x}_2 \in \Omega_k} \norm{\bm{x}_1 -\bm{x}_2}_2 \leq h$, $\Omega \Delta \Omega^N:=(\Omega \setminus \Omega^N)\cup (\Omega^N \setminus \Omega) = \Omega \ \setminus \Omega^N$ since $\Omega^N \subseteq \Omega$, and $|\Omega \Delta \Omega^N|$ is its area.
    \label{thm:integral_error_estimate}
\end{theorem}

\begin{remark}
    The estimate above highlights several important aspects. First, in regions where the function is sufficiently smooth, rapid convergence is observed. In contrast, regions with discontinuities exhibit significantly slower convergence due to the inherent difficulty of approximating abrupt changes. 
    
    This issue could arise, for example, when approximating a function defined on a domain $\Omega$ using basis functions supported on a larger domain $\widetilde{\Omega}$, where $\Omega \subset \widetilde{\Omega}$. In particular, Theorem~\ref{lemma:general_coeff_error_estimate} shows that part of the projection error stems from approximating the inner products \(\braket{f}{\phi_i}_{L_2(\widetilde{\Omega})}\). If \(f\) is only defined on \(\Omega\), this inner product is typically computed by extending \(f\) to zero outside \(\Omega\), which might introduce a discontinuity at the boundary of \(\Omega\). This discontinuity makes accurate integration more difficult near the boundary of $\Omega$. 
    
    Finally, approximating the domain itself introduces an additional source of error.
\end{remark}

The previous error estimate requires a mesh. To deal with point clouds, one can consider Monte Carlo sampling to approximate the integral, assume the samples are sampled according to some specific distribution, and obtain a probabilistic bound with an error of $\mathcal{O}(1/\sqrt{N})$ \cite{ross2022simulation}. To remove the randomness assumption and get a non-probabilistic bound, one can employ quasi-Monte Carlo schemes. Instead of randomly sampling and approximating the integral in that way, one uses a sum $\frac{1}{N}\sum_{i=1}^N f(\bm{x}_i)$ with deterministic $\bm{x}_i \in \Omega$ to approximate $\int_\Omega f(\bm{x}) \dint \bm{x}$. In this case, one can employ Koksma-Hlawka type inequalities to bound the approximation error \cite{niederreiter1992random, kuipers2012uniform, pausinger2015koksma}. The general form of such inequalities bounds the error as a product of two terms: 
\begin{itemize}
    \item a discrepancy term that measures how well the sampled  points are approximating the given distribution or measure, and 
    \item a measure of variation of the integrand.
\end{itemize}
In particular, Theorem 4.3 of \cite{pausinger2015koksma} proves a Koksma-Hlawka inequality when integrating over general compact domains. For completeness, the complete statement of the theorem is provided in \ref{SM:koksma-hlawka_ineq}. 

\subsection{Neural ODE flow operator} \label{sec:node_and_decoder}
After encoding the input into an initial latent code $\mathbf{z}(0)= \mathbf{z}_0 = E(u_0)$, the neural ODE in Equation \eqref{eq:latent_ode} returns a latent trajectory $\mathbf{z}(t)$. Any numerical solver can be used; however, it only yields values at the points $\{t_i\}_{i=0}^{N_{\delta_t}-1}$ that are used to solve the differential equation. It is common to employ an interpolation method to obtain latent vectors at intermediate time points, which is necessary for evaluating the neural operator at arbitrary values of $t$.  

Given a time discretization \( \bm{t} := \{t_i\}_{i=0}^{N_{\delta_t}-1} \subset [0, T] \) with $N_{\delta_t}$ time instances, a numerical solver \( \Psi(\mathbf{z} ; \bm{t}) \) can be used to approximate the discrete solution at times $t_i$,
\[
\Psi(\mathbf{z} ; \bm{t}) = \{\widehat{\mathbf{z}}(t_i)\}_{i=0}^{N_{\delta_t}-1}, \quad \text{with } \widehat{\mathbf{z}}(t_0) = \mathbf{z},
\]
where $\widehat{\mathbf{z}}(t_i)$ denotes the approximated discrete solution of the solver at time instance $t_i$ and $\mathbf{z} \in \R^{d_z} $ denotes the initial condition. To recover a continuous representation over time from the discrete solutions, Hermite spline interpolations can be applied. In the scalar case, Hermite interpolation constructs a polynomial \( s(t) \) of degree \( 2p - 1 \) that approximates a general smooth function \( f: [0, h] \to \mathbb{R} \) over a small interval, with the following conditions on the polynomial $s$,
\[
\frac{\mathrm{d}^{i}}{\mathrm{d}x^{i}}s(0) = \frac{\mathrm{d}^{i}}{\mathrm{d}x^{i}}f(0), \quad \frac{\mathrm{d}^{i}}{\mathrm{d}x^{i}}s(h) = \frac{\mathrm{d}^{i}}{\mathrm{d}x^{i}} f(h), \quad\quad \text{for } i = 0, \dots, p - 1,
\]
Since \( \widehat{\mathbf{z}}(t_i) \in \R^{d_z} \), Hermite interpolation can be applied component-wise on each interval $[t_i, t_{i+1}]$. For the $i$-th interval $[t_i, t_{i+1}]$ and $j$th coordinate \( j = 1, \ldots, d_z \), a univariate Hermite interpolant \( s_{ij}(t) \) can be constructed. The full interpolated value for $t \in [t_i,t_{i+1}]$ ($i \in \{ 0,\dots,N_{\delta_t}-2\}$) can be written as $\bm{s}(t) \coloneqq \bm{s}_i(t)$ with $\bm{s}_i(t) = [s_{i1}(t), s_{i2}(t), \cdots, s_{id_z}(t)]^{\top}$.  

For demonstration, cubic interpolation is assumed here, namely $p=2$. The cubic interpolation uses \( \{\widehat{\mathbf{z}}(t_i)\}_{i=0}^{N_{\delta_t}-1} \) and the exact time derivatives $\mathbf{v}(\widehat{\mathbf{z}}(t_i), t)$. Combining the numerical solver with the interpolation scheme yields a time-continuous approximation \( \widehat{\mathcal{F}}_t(\mathbf{z}; \bm{t}) \) of the exact flow map \( \mathcal{F}_t(\mathbf{z}) \). In particular, denoting the interpolation as $\bm{s}(t ; \Psi(\mathbf{z} ; \bm{t}), \bm{t})$ to highlight its dependency on the time-discretization and the numerical solver, the time-continuous approximation is
\begin{equation}
    \widehat{\mathcal{F}}_t(\mathbf{z}; \bm{t}) \coloneqq \bm{s}(t ; \Psi(\mathbf{z} ; \bm{t}), \bm{t}).
    \label{eq:full_numerical_solver_ODE}
\end{equation}
The choice of numerical solver and the order of the spline interpolation method directly determine the error between $\widehat{\mathcal{F}}_t$ and $\mathcal{F}_t$. While this holds for any $p \geq 1$ (as shown in \ref{SM:p_unequal_to_2_case_error_estimate_NODE}), we demonstrate the case when $p=2$. 
\begin{theorem}
    Assume we have a numerical ODE solver $\Psi$ with global error order $\mathcal{O}(\delta_t^q)$ and a time discretization $\bm{t} := \{t_i\}_{i=0}^{N_{\delta_t}-1}$ satisfying $\sup_{i\in\{1, \ldots, N_{\delta_t}-1\}}|t_i - t_{i-1}| \leq \delta_t$. Furthermore, assume $\mathbf{v}$ is $L_v$-Lipschitz in $\mathbf{z}$ and that for $j=1, \ldots, d_z$ the functions $(\mathbf{z}, t) \mapsto (R_3(\mathbf{z}, t))_j$ are in $L^\infty(Z\times \R)$ for $R_{0}(\mathbf{z}, t) \coloneqq \mathbf{v}(\mathbf{z}, t)$ and $R_k$ defined as: 
    \begin{equation*}
        \begin{aligned}
             R_{k}(\mathbf{z}, t) \coloneqq \left(\sum_{j=0}^{d_z} \sum_{l=0}^{k-1} {k-1 \choose l} 
            \left(\frac{\partial}{\partial \mathrm{z}_j} R_{k-1-l}(\mathbf{z}, t)\right) (R_{l}(\mathbf{z}, t))_j\right) + \frac{\partial}{\partial t} R_{k-1}(\mathbf{z}, t).
        \end{aligned}
    \end{equation*}
    Then, when using cubic Hermite interpolation in Equation \eqref{eq:full_numerical_solver_ODE}, the following error estimate applies,
    \begin{equation*}
        \norm{\mathcal{F}_t(\mathbf{z}) - \widehat{\mathcal{F}}_t(\mathbf{z};\bm{t})}_2 = \mathcal{O}\left(\delta_t^{\min(4,q)}\right).
    \end{equation*}
    \label{thm:error_estimate_ode_discretization_p=2}
\end{theorem}
\begin{proof}
    Take a time $t \in [t_i, t_{i+1}]$. Then $\mathcal{F}_t(\mathbf{z}):=\mathbf{z}(t)$ and $\widehat{\mathcal{F}}_t(\mathbf{z} ; \bm{t}) = \bm{s}_i(t)$. Define the numerical approximation $\widehat{\mathbf{z}}(\bm{t}):=\{\widehat{\mathbf{z}}(t_i)\}_{i=0}^{N_{\delta_t}-1}$ to Equation \eqref{eq:latent_ode}. Moreover, let $\tilde{\bm{s}}_i$ be defined analogously to $\bm{s}_i$, but constructed by interpolating the ground truth values $\mathbf{z}(t_i)$ and $\mathbf{z}(t_{i+1})$. Then for $t \in [t_i, t_{i+1}]$ it follows that,
    \begin{equation}
        \begin{split}
            \norm{\mathbf{z}(t) - \bm{s}_i(t)}_2 & \leq \norm{\mathbf{z}(t) - \tilde{\bm{s}}_i(t)}_2 + \norm{\tilde{\bm{s}}_i(t) - \bm{s}_i(t)}_2 \\
            & \leq \frac{\delta_t^{4}}{2^{4}(4)!} \left({\sum_{j=1}^{d_z} \norm{\frac{\mathrm{d}^{4}}{\mathrm{d}t^{4}} \mathrm{z}_j}_{L^\infty(t_i,t_{i+1})}^2} \right)^{\frac{1}{2}} + \norm{\tilde{\bm{s}}_i(t) - \bm{s}_i(t)}_2,
        \end{split}
        \label{eq:general_error_estimate_ode_solver}
    \end{equation} 
    where the final inequality follows from standard Hermite interpolation bounds \cite{stoer1980introduction, corless2013graduate}. To show the existence and boundedness of the fourth-order derivative, observe that,
    \begin{equation*}
        \frac{\mathrm{d}^{k}}{\mathrm{d}t^{k}} \mathrm{z}_i  = \frac{\mathrm{d}^{k-1}}{\mathrm{d}\mathrm{t}^{k-1}} v_i(\mathbf{z}(t), t) = \frac{\mathrm{d}^{k-2}}{\mathrm{d}t^{k-2}} \left(\sum_{j=0}^{d_z} \frac{\partial v_i(\mathbf{z}(t), t)}{\partial \mathrm{z}_j} \frac{\mathrm{d}\mathrm{z}_j}{\mathrm{d}t} + \frac{\partial v_i(\mathbf{z}(t), t)}{\partial t}\right),
    \end{equation*}
    which can be rewritten using the general Leibniz rule to,
    \begin{equation*}
        \left(\sum_{j=0}^{d_z} \sum_{l=0}^{k-2} {k-2 \choose l} 
            \left(\frac{\partial}{\partial \mathrm{z}_j}  \frac{\mathrm{d}^{k-2 - l}}{\mathrm{d}t^{k-2 - l}}(v_i(\mathbf{z}(t), t))\right) \frac{\mathrm{d}^{l}}{\mathrm{d}t^{l}} v_j(\mathbf{z}(t), t)\right) + \frac{\partial}{\partial t} \frac{\mathrm{d}^{k-2}}{\mathrm{d}t^{k-2}}(v_i(\mathbf{z}(t), t).
    \end{equation*}
    Hence, $\frac{\mathrm{d}^{k}}{\mathrm{d}t^{k}} \mathrm{z}_j = (R_{k-1}(\mathbf{z}(t), t))_j$, which implies $\norm{\frac{\mathrm{d}^{4}}{\mathrm{d}t^{4}} \mathrm{z}_j}_{L^\infty(t_i,t_{i+1})}^2 < \infty$ as $(\mathbf{z}, t) \mapsto (R_3(\mathbf{z}, t))_j$ is a function in $L^\infty(Z\times \R)$.

    By defining $\delta_i:=(t_{i+1}-t_i)$, $h_{00}(x)\coloneqq (2 x^3 - 3x^2 +1)$, $h_{10}(x)\coloneqq (x^3 - 2x^2 + x)$, $h_{01}(x)\coloneqq (-2 x^3 + 3x^2)$, $h_{11}(x)\coloneqq (x^3 - x^2)$, and $\tilde{t}\coloneqq \frac{t - t_i}{\delta_i}$, the cubic spline interpolation is given by $\tilde{\bm{s}}_i(t) = h_{00}(\tilde{t}) \mathbf{z}(t_i) + h_{01}(\tilde{t}) \mathbf{z}(t_{i+1}) + h_{10}(\tilde{t}) \delta_i \frac{\mathrm{d} \mathbf{z}(t_i)}{\mathrm{d}t} + h_{11}(\tilde{t}) \delta_i \frac{\mathrm{d} \mathbf{z}(t_{i+1})}{\mathrm{d}t}$. With this formulation for the spline interpolation and defining $\mathbf{e}(t)\coloneqq \mathbf{z}(t) - \widehat{\mathbf{z}}(t)$, the interpolation error on each interval can be written as,
    \begin{equation*}
        \begin{split}
            \norm{\tilde{\bm{s}}_{i}(t) - \bm{s}_{i}(t)}_2 & = \norm{\sum_{k=0}^1 h_{0k}(\tilde{t}) \mathbf{e}(t_{i+k}) +h_{1k}(\tilde{t}) \delta_i \left(\frac{\mathrm{d} }{\mathrm{d}t} \mathbf{e}(t_{i+k})\right)}_2 \\
            & \leq  \sum_{k=0}^1 |h_{0k}(\tilde{t})| \norm{\mathbf{e}(t_{i+k})}_2 + |h_{1k}(\tilde{t}) |\delta_i \norm{\frac{\mathrm{d} }{\mathrm{d}t} \mathbf{e}(t_{i+k})}_2.
        \end{split} 
    \end{equation*}
    For $j = i, i+1$, $\norm{\frac{\mathrm{d}}{\mathrm{d}t}\mathbf{e}(t_j)}_2 = \norm{\frac{\mathrm{d} \mathbf{z}(t_j)}{\mathrm{d}t} - \frac{\mathrm{d} \widehat{\mathbf{z}}(t_j)}{\mathrm{d}t}}_2  = \norm{\mathbf{v}(\mathbf{z}(t_j), t_j) - \mathbf{v}(\widehat{\mathbf{z}}(t_j), t_j)}_2$, and hence $\norm{\frac{\mathrm{d}}{\mathrm{d}t}\mathbf{e}(t_j)}_2 \leq L_v \norm{\mathbf{z}(t_j) - \widehat{\mathbf{z}}(t_j)} = L_v \norm{\mathbf{e}(t_j)}$ by Lipschitz-continuity of $\mathbf{v}$. Combining this result with the preceding inequality yields,
    \begin{equation*}
        \begin{split}
            \norm{\tilde{\bm{s}}_{i}(t) - \bm{s}_{i}(t)}_2 & \leq (|h_{00}(\tilde{t})| + L_v|h_{10}(\tilde{t})|\delta_i) \norm{\mathbf{z}(t_i) - \widehat{\mathbf{z}}(t_i)}_2 \\
            & + (|h_{01}(\tilde{t})| + L_v|h_{11}(\tilde{t})|\delta_i) \norm{\mathbf{z}(t_{i+1}) - \widehat{\mathbf{z}}(t_{i+1})}_2 \\
            & \leq \left(|h_{00}(\tilde{t})| + |h_{01}(\tilde{t})|+ L_v(|h_{10}(\tilde{t})| + |h_{11}(\tilde{t})|)\delta_t \right) C \delta_t^q,
        \end{split} 
    \end{equation*}
    where the last inequality holds as the numerical solver is of the order $q$ and $\delta_i \leq \delta_t$. By integrating this result with the inequality shown in \eqref{eq:general_error_estimate_ode_solver} and the boundedness of $h_{kl}$ on $[0,1]$, we get an error of order $\delta_t^{\min(4,q)}$, confirming the claim.
\end{proof}
\begin{remark}
    The $\mathbf{v}$ in Equation \eqref{eq:latent_ode}, and used in Theorem \ref{thm:error_estimate_ode_discretization_p=2}, is parameterized by a neural network, which is trained jointly with the encoder and decoder. In the reduced basis method and Galerkin projection methods, the $\mathbf{v}$ is not learned. In particular, only the encoder and decoder are learned before solving a reduced system with $\mathbf{v}$ defined via an optimization problem. We can take a similar strategy in our setup. Particularly, given a current estimate $\mathbf{z}(t)$ and noting that $\frac{\partial}{\partial t}D(\mathbf{z}(t))(\bm{x}) = \braket{\nabla_z D(\mathbf{z})(\bm{x})}{\frac{\mathrm{d}}{\mathrm{d} t} \mathbf{z}(t)}$, we can choose the velocity vector field that corresponds to the best projection of $\mathcal{N}(D(\mathbf{z}(t)), \cdot)$ onto the tangent space of our latent manifold,
    \begin{equation}
        \min_{\dot{\mathbf{z}} := \frac{\mathrm{d}}{\mathrm{d} t}\bm{\mathbf{z}}(t)} \int_\Omega \norm{\braket{\nabla_z D(\mathbf{z})(\bm{x})}{\dot{\mathbf{z}}} - \mathcal{N}(D(\mathbf{z}(t)), \bm{x})}^2 \dint \bm{x}.
        \label{eq:best_tangent_proj}
    \end{equation}     
    The decoder $D \colon X \times Z \to \mathbb{R}$ is trained in advance and kept fixed during optimization. This contrasts with neural Galerkin methods \cite{bruna2024neural} and evolutionary deep neural networks \cite{kast2024positional}, which eliminate the latent space $Z$ and work directly with network parameters. Classical Galerkin methods do incorporate a latent representation and use a linear decoder in \eqref{eq:best_tangent_proj} that maps latent coefficients to a linear subspace. Manifold Galerkin \cite{lee2020model} relaxes this linearity by employing a neural network decoder, but is restricted to ODEs and therefore replaces the $L_2(\Omega)$ norm with an $\ell_2$ norm.
    
    Since the decoder in \eqref{eq:best_tangent_proj} maps latent variables to functions, the resulting projection can be interpreted as a manifold Galerkin method for PDEs that unifies the manifold and neural Galerkin viewpoints. To connect this formulation with Theorem~\ref{thm:error_estimate_ode_discretization_p=2}, fix $\bm{x}$ and let $J_D(\mathbf{z},\bm{x})$ denote the Jacobian of $D(\mathbf{z})(\bm{x})$ with respect to $\mathbf{z}$. The solution of \eqref{eq:best_tangent_proj} is then given by the following matrix equation:
    \begin{equation*}
        \left(\int_\Omega J_D (\mathbf{z}(t), \bm{x}) ^T J_D(\mathbf{z}(t), \bm{x}) \dint \bm{x}\right) \dot{\mathbf{z}} =  \left( \int_\Omega J_D(\mathbf{z}, \bm{x}) \mathcal{N}(D(\mathbf{z}(t)), \bm{x}) \dint \bm{x}\right).
    \end{equation*}
    To apply Theorem \ref{thm:error_estimate_ode_discretization_p=2}, it is necessary to show that $(\mathbf{z}, t) \mapsto (R_3(\mathbf{z}, t))_j$ is in $L^\infty(Z\times \R)$ for $j=1, \ldots, d_z$ and that the solutions to the above system of equations are Lipschitz in $\mathbf{z}(t)$. The latter property parallels the result established in Theorem \ref{lemma:general_coeff_error_estimate}. Overall, under appropriate assumptions on the decoder, Theorem \ref{thm:error_estimate_ode_discretization_p=2} remains valid. 
\end{remark}

\begin{remark}
    Theorem \ref{thm:error_estimate_ode_discretization_p=2} assumes that $\mathbf{v}$ is Lipschitz and that the components of $R_3$, as defined in Theorem \ref{thm:error_estimate_ode_discretization_p=2}, belong to $L^\infty(Z \times R)$. In \ref{SM:lipschitz_cont_of_NNs_and_RONOM}, we show that neural networks with common activation functions are Lipschitz, and hence so is $\mathbf{v}$. Moreover, \ref{SM:bounded_derivs_NNs} shows that the derivatives of neural networks with common activation functions are bounded. Since the definition of $R_k$ is recursive and depends only polynomially on $\mathbf{v}$ and its higher order derivatives, as can be seen by unrolling the recursion, it follows that $R_k$ is automatically bounded. Therefore, the $L^\infty(Z \times R)$ assumption is not restrictive. Of course, depending on the learned structure and the system under consideration, the $L^\infty(Z \times R)$ norms of the components of $R_k$ may still be large.
\end{remark}

\subsection{Decoder}\label{sec:decoder}
The decoder maps latent vectors $\mathbf{z}(t)$ to functions. A typical example is the DeepONet, where $\mathbf{z}(t)$ feeds into the branch net and spatial coordinates into the basis functions. To align with standard ROM methods, a DeepONet-style decoder is employed, utilizing fixed, non-learnable basis functions. This structure closely relates to the POD-DeepONet \cite{lu2022comprehensive}, which chooses the basis a-priori via POD.

In particular, our decoder is based on the optimal recovery method from Batlle et al.\ \cite{batlle2024kernel}. Let $\{\bm{x}_i\}_{i=1}^{N_c} \subset \Omega$ be a set of distinct spatial locations, and let $K\colon \Omega \times \Omega \to \mathbb{R}$ be a positive definite kernel associated to a reproducing kernel Hilbert space $\mathcal{H}_K$ with norm $\|\cdot\|_{K}$. The regularized version of the optimal recovery problem seeks the function $u$ that solves:
\begin{equation*}
\begin{aligned}
\min_{u \in \mathcal{H}_K} \quad & \sum_{i=1}^{N_c} |u(\bm{x}_i) - y_i|^2 + \eta \|u\|_K ^2,
\end{aligned}
\end{equation*}
where $\eta \rightarrow \infty$ corresponds to the optimal recovery problem, which finds $u$ of minimal RKHS norm that interpolates the given data. Let the coefficients $\{a_i\}_{i=1}^{N_c}$ originate from the linear system $K_\eta\bm{a} = \mathbf{y}$, where $K_\eta \coloneqq K(\mathbf{X}, \mathbf{X}) + \eta I$, $\bm{a} = [a_1, \dots, a_{N_c}]^\top$ and $(\mathbf{X},\mathbf{y}) = \{(\bm{x}_i,y_i)\}_{i=1}^{N_c}$. The solution to the regularized problem can be written as,
\[
u\left(\bm{x}\,\big|\,\{(\bm{x}_i, y_i)\}_{i=1}^{N_c}\right) = \sum_{i=1}^{N_c} a_i K(\bm{x}, \bm{x}_i) = K(\bm{x}, \mathbf{X}) K_\eta^{-1} \mathbf{y} .
\]

When the solution operator to the PDE is trained, the values $y_i$ are unknown. Consequently, a map $D_d\colon Z \to \mathbb{R}^{N_c}$ is defined from the latent space to the predicted values $y_i$. The complete form of the decoder is then expressed as,
\begin{equation}
D(\mathbf{z})(\bm{x}) = \sum_{i=1}^{N_b} \left(K_\eta^{-1} D_d(\mathbf{z}) \right)_i K(\bm{x}, \bm{x}_i) =  K(\bm{x}, \mathbf{X}) K_\eta^{-1} D_d(\mathbf{z}).
\label{eq:decoder}
\end{equation}
An important feature of this construction is that $D_d$ can be any latent-to-vector map used in the ROM literature. The decoder outputs a function via upsampling with a kernel, and its Lipschitz continuity can be analyzed straightforwardly for any norm on functions.
\begin{theorem}
    Let $\norm{\cdot}_\U$ be a norm  on functions and $D_d$ be $L_{D_d}$-Lipschitz. Then $D$ in Equation \eqref{eq:decoder} is $L_D$-Lipschitz in $\norm{\cdot}_\U$ where $L_D := L_{D_d} C_F \sqrt{\norm{K_\eta^{-T}A K_\eta^{-1}}_{2}}$ and $A$ is defined by:
    \begin{equation*}
        A_{ij} = \begin{cases}
            \braket{K(\cdot, \bm{x}_i)}{K(\cdot, \bm{x}_j)}_\U, \quad \norm{\cdot}_\U \text{ comes from an inner product} \\
            K(\bm{x}_i, \bm{x}_j), \quad \, \, \qquad \qquad \text{otherwise}
        \end{cases}
    \end{equation*}
    and $C_F = 1$ when $\norm{\cdot}_\U$ comes from an inner product and depends on $\norm{\cdot}_\U$ and $K$ otherwise. 
    \label{thm:D_lipschitzness}
\end{theorem}
\begin{proof}
    For the case that an inner product does not induce the norm, we can define $\braket{\sum_{i=1}^{N_c}a_i K(\cdot, \bm{x}_i)}{\sum_{i=1}^{N_c}b_j K(\cdot, \bm{x}_j)}_{\H_K}:= \sum_{i,j=1}^{N_c}a_i K(\bm{x}_i, \bm{x}_j) b_j$ as the inner product on our space $\mathSpan(K(\cdot, \bm{x}_i) \mid i=1, \ldots, N_c)$. By equivalence of norms on finite-dimensional spaces, $\norm{D(\mathbf{z}_1) - D(\mathbf{z}_2)}_\U \leq \widetilde{C}_\U \norm{D(\mathbf{z}_1) - D(\mathbf{z}_2)}_{\H_K}$ for some constant $\widetilde{C}_\U$. When considering the $\H_K$ inner product or when considering a norm $\norm{\cdot}_\U$ that comes from an inner product, Equation \eqref{eq:decoder} gives us that $\norm{D(\mathbf{z}_1) - D(\mathbf{z}_2)}^2$ can be rewritten to,
    \begin{align*}
             & \braket{\sum_{i=1}^{N_c} (K_\eta^{-1} (D_d(\mathbf{z}_1) - D_d(\mathbf{z}_2)))_i K(\cdot, \bm{x}_i)}{\sum_{j=1}^{N_c} (K_\eta^{-1} (D_d(\mathbf{z}_1) - D_d(\mathbf{z}_2)))_j K(\cdot, \bm{x}_j)} \\
            & = \sum_{i,j =1}^{N_c} (K_\eta^{-1} (D_d(\mathbf{z}_1) - D_d(\mathbf{z}_2)))_i \braket{K(\cdot, \bm{x}_i)}{K(\cdot, \bm{x}_j)} (K_\eta^{-1} (D_d(\mathbf{z}_1) - D_d(\mathbf{z}_2)))_j,
    \end{align*}
    where the inner product is $\braket{\cdot}{\cdot}_\U$ when $\norm{\cdot}_\U$ comes from an inner product or $\braket{\cdot}{\cdot}_{\H_K}$ otherwise. Let $C_F = 1$ when $\norm{\cdot}_\U$ comes from an inner product and let $C_F = \widetilde{C}_\U$ otherwise. Combining the previous observations with the definition of the matrix $A$, it leads to
    \begin{align*}
        \norm{D(\mathbf{z}_1) - D(\mathbf{z}_2)}_\U^2 & \leq C_F^2 (K_\eta^{-1} (D_d(\mathbf{z}_1) - D_d(\mathbf{z}_2)))^{\top} A (K_\eta^{-1} (D_d(\mathbf{z}_1) - D_d(\mathbf{z}_2))) \\
        & = C_F^2 (D_d(\mathbf{z}_1) - D_d(\mathbf{z}_2))^{\top} \left( K_\eta^{-\top}  A K_\eta^{-1} \right) (D_d(\mathbf{z}_1) - D_d(\mathbf{z}_2)) \\
        & \leq C_F^2 \norm{K_\eta^{-\top}  A K_\eta^{-1}}_2 \norm{D(\mathbf{z}_1) - D(\mathbf{z}_2)}_2^2 \\
        & \leq L_{D_d}^2 C_F^2 \norm{K_\eta^{-\top}  A K_\eta^{-1}}_2 \norm{\mathbf{z}_1 - \mathbf{z}_2}_2^2  = L_D^2 \norm{\mathbf{z}_1 - \mathbf{z}_2}_2^2.
    \end{align*}
\end{proof}

\section{RONOM error estimates}
Discretization plays an essential role in both operator learning and ROM. In the neural operator literature, discretization convergence and discretization invariance characterize whether discretizations of a learned infinite-dimensional operator converge to this infinite-dimensional operator. In the ROM literature, similar questions arise when analyzing the discretization error of the reduced system. We present a theorem that, for the first time, bridges these questions in neural operator learning and ROM. Before stating this result, we provide a general a-posteriori error estimate for the discrepancy between our infinite-dimensional reconstruction and the ground truth solution of the PDE.

\subsection{An a-posteriori error estimate}
The following theorem is closely related to Proposition 1 from Farenga et al.\ \cite{farenga2025latent}. Whereas their result is formulated for ODEs, our theorem is established in the context of PDEs. In addition, our error estimate is expressed solely in terms of quantities that can be computed from the model, in contrast to theirs, which also depends on the unknown ground truth solution. 
\begin{theorem}\label{thm:a-posteriori_err_estimate}
     Assume the strong solution $u(t)$ to the PDE in Equation \eqref{eq:pde} exists and let $\hat{u}(t) := D(\mathbf{z}(t))$ be our approximation with $\frac{\mathrm{d}}{\mathrm{d}t}\mathbf{z}(t) = \mathbf{v}(\mathbf{z}(t), t)$ with $\mathbf{z}(0)=\mathbf{z}_0$. Furthermore, suppose a Hilbert space $\U$ is given such that $u(t), \hat{u}(t) \in \U$ and $\mathcal{N}(u, \cdot) \in \U \, \, \forall u \in \U$. Assuming the nonlinear operator $\mathcal{N}$ is $L_\mathcal{N}$-Lipschitz with respect to $\norm{\cdot}_\U$, the error $e(t):=\norm{u(t) - \hat{u}(t)}_\U$ satisfies,
    \begin{equation*}
        e(t) \leq \mathrm{e}^{L_\mathcal{N} t}\left(e(0) + \int_0^t\left( \norm{\mathcal{N}(\hat{u}(t), \cdot)  - \frac{\partial}{\partial t}\hat{u}(t)}_\U \right) \mathrm{d}t \right).
    \end{equation*}
\end{theorem}

\begin{proof}
    Note that $\frac{\mathrm{d}}{\mathrm{d}t} \left(\frac{1}{2}e(t)^2 \right) = \braket{u(t) - \hat{u}(t)}{\frac{\partial}{\partial t}(u(t) - \hat{u}(t))}_\U$ and:
    \begin{align*}
            & \braket{u(t) - \hat{u}(t)}{\frac{\partial}{\partial t}(u(t) - \hat{u}(t))}_\U = \braket{u(t) - \hat{u}(t)}{\mathcal{N}(u(t), \cdot)  - \frac{\partial}{\partial t}\hat{u}(t)}_\U \\
            & = \braket{u(t) - \hat{u}(t)}{\mathcal{N}(u(t), \cdot)  - \mathcal{N}(\hat{u}(t), \cdot)}_\U +  \braket{u(t) - \hat{u}(t)}{\mathcal{N}(\hat{u}(t), \cdot) -  \frac{\partial}{\partial t}\hat{u}(t)}_\U \\
            & \leq L_\mathcal{N} \norm{u(t) - \hat{u}(t)}_\U^2 + \norm{\mathcal{N}(\hat{u}(t), \cdot)  - \frac{\partial}{\partial t}\hat{u}(t)}_\U \norm{u(t) - \hat{u}(t)}_\U \\
            & = L_\mathcal{N} e(t)^2 + \norm{\mathcal{N}(\hat{u}(t), \cdot)  - \frac{\partial}{\partial t}\hat{u}(t)}_\U e(t).
    \end{align*}
    The inequality for the first term arises from the fact that the differential operator $\mathcal{N}$ is $L_\mathcal{N}$-Lipschitz, and the second term follows from the Cauchy–Schwarz inequality. Using $e(t) \frac{\mathrm{d}}{\mathrm{d}t} e(t) = \frac{\mathrm{d}}{\mathrm{d} t} \left(\frac{1}{2}e(t)^2 \right)$ and dividing both sides by $e(t)$ then yields,
    \begin{equation*}
        \frac{\mathrm{d}}{\mathrm{d}t} e(t) \leq L_\mathcal{N} e(t) + \norm{\mathcal{N}(\hat{u}(t), \cdot)   - \frac{\partial}{\partial t}\hat{u}(t)}_\U.
    \end{equation*}
    Using Grönwall's inequality yields the desired inequality.
\end{proof}

 The inequality highlights two key aspects. It shows that errors in encoding the initial condition propagate through $e(0)$. If the initial condition is reconstructed perfectly, the remaining error stems from $\frac{\partial}{\partial t} \hat{u}(t)$ not precisely following the evolution dictated by the PDE.

 \begin{remark}
     The above theorem assumes Lipschitz continuity of $\mathcal{N}$, which is not satisfied by all PDEs. Nevertheless, this assumption is reasonable in our setting. With only a finite number of training samples, the system can effectively be restricted to a manifold $\mathcal{M}$ within a compact set. This is further supported by the fact that traditional ROM latent space approaches often assume many systems evolve near lower-dimensional manifolds. For a continuous operator, Lipschitz continuity on such compact sets naturally follows from local Lipschitzness. 

    Another advantage of assuming Lipschitzness is that it enables the application of Theorem~\ref{thm:a-posteriori_err_estimate}. Hence, even though the Lipschitz assumption does not hold for all PDEs, it allows Theorem~\ref{thm:a-posteriori_err_estimate} to provide valuable intuition on approximation errors through an a-posteriori estimate.    
 \end{remark}

\subsection{Discretization error estimate} \label{sec:discretization_error_estimate}
As mentioned in the introduction, neural operator frameworks are frequently developed without a thorough analysis of how accurately their discretized implementations approximate the underlying infinite-dimensional operators. In contrast, ROM techniques, especially reduced basis methods, are supported by rigorous error analyses, which provide theoretical guidance on the accuracy of the numerical construction and approximation of the reduced system for time-dependent problems.

In the theorem below, we unify these perspectives by providing an error estimate for the full RONOM pipeline, which also establishes discretization convergence. 

\begin{theorem}[Discretization error estimate RONOM]\label{thm:discretization_error_estimate_RONOM}
    Assume that $\mathcal{E}_\varphi$ in the encoder is $L_{\mathcal{E}_\varphi}$-Lipschitz, the decoder $D \colon Z \to \mathcal{U} \subset L_2(\Omega)$ is $L_D$-Lipschitz for some Banach Space $\U$ (e.g., see Theorem \ref{thm:D_lipschitzness}), and the assumptions of Theorem \ref{thm:error_estimate_ode_discretization_p=2} are satisfied. Let $\widehat{\mathcal{K}}^\dagger(U_0, t)\coloneqq  (D \circ \widehat{\mathcal{F}}_t \circ \widehat{E})(U_0)$ approximate $\mathcal{K}^\dagger(u_0, t)(\bm{x}) := (D \circ \mathcal{F}_t \circ E)(u_0)(\bm{x})$, where $U_0 \coloneqq \{\bm{x}_i, u_{0i}\}_{i=1}^n$, $\widehat{\mathcal{F}}_t(\mathbf{z}; \bm{t})$ is given in Equation \eqref{eq:full_numerical_solver_ODE} and the discrete encoder $\widehat{E}$ results from approximating $\P_\V^\lambda$ using $\widehat{\mathcal{P}}_\V^\lambda$, i.e.,
    \begin{equation*}
        \widehat{E}(U_0) \coloneqq (\mathcal{E}_\varphi \circ M \circ \widehat{\mathcal{P}}_\V^\lambda)(U_0).
        \label{eq:discretized_encoder}
    \end{equation*}
    Then with $q$ the order of the numerical ODE solver, $L_v$ the Lipschitz constant of $\mathbf{v}$, time discretization $\bm{t}:=\{t_i\}_{i=0}^{N_{\delta_t}-1}$ satisfying $\sup_{i=1, \ldots, N_{\delta_t}-1} |t_i - t_{i-1}| \leq\delta_t$, $\mathrm{M}_\phi$ the matrix with columns $M \phi_i$, and $L(t) \coloneqq L_D e^{L_v t} \norm{\mathrm{M}_\phi}_2$, the following holds,
    \begin{equation*} 
            \norm{ \mathcal{K}^\dagger(u_0, t) - \widehat{\mathcal{K}}^\dagger(U_0, t)}_\U \leq L(t) e_a(u_0, U_0)  + L_D\mathcal{O}\left(\delta_t^{\min(4, q)}\right) , 
    \end{equation*}
    where $e_a(u_0, U_0)$ denotes the bound from Theorem \ref{lemma:general_coeff_error_estimate},
    \begin{equation*}
    \begin{aligned}
        e_a(u_0, U_0) &\coloneqq \frac{ \norm{\mathrm{L}^{-1}}_2}{\lambda}\Big(\norm{\braket{u_0}{\bm{\phi}}_{L_2(\Omega)} - \Tilde{\Phi} D_w \bm{u_0}}_2 \\
        &+ \norm{(\Tilde{\Phi} D_w \Tilde{\Phi}^T - \Phi)}_2 \norm{(\Phi + \lambda \mathrm{L})^{-1}}_2 \norm{\braket{u_0}{\bm{\phi}}_{L_2(\Omega)}}_2 \Big).
    \end{aligned}
    \end{equation*}
\end{theorem}

\begin{proof}
    An error estimate for the discretized encoder is first established. This result is then combined with previously derived estimates to obtain a comprehensive error bound for the overall framework.
    \subsubsection*{Error estimate between $E$ and $\widehat{E}$}
    Let $a^\dagger(u_0)$ and $a(U_0)$ be the coefficients such that $\P_\V^\lambda(u_0) = \sum_{i=1}^{N_b} a^\dagger_i(u_0) \phi_i$ and $\widehat{\P}_\V^\lambda(U_0) = \sum_{i=1}^{N_b} a_i(U_0) \phi_i$, respectively. Owing to linearity of $M$ and $L_{\mathcal{E}_\varphi}$ Lipschitzness of $\mathcal{E}_\varphi$, we obtain the following estimate:
    \begin{equation*}
        \begin{aligned}
            \norm{E(u_0) - \widehat{E}(U_0)}_2 & = \norm{\mathcal{E}_\varphi \circ M\circ\mathcal{P}_\mathcal{V}^\lambda(u_0) - \mathcal{E}_\varphi \circ M\circ \widehat{\mathcal{P}}_\mathcal{V}^\lambda(U_0)}_2 \\
            & \leq L_{\mathcal{E}_\varphi} \norm{\left(\sum_{i=1}^{N_b} (a_i^\dagger(u_0) - a_i(U_0))M\phi_i\right)}_2 \\
            & = L_{\mathcal{E}_\varphi} \norm{\mathrm{M}_\phi (a^\dagger(u_0) - a(U_0))}_2 \\
            & \leq L_{\mathcal{E}_\varphi} \norm{\mathrm{M}_\phi}_2 \norm{a^\dagger(u_0) - a(U_0)}_2.
        \end{aligned}
    \end{equation*}

    \subsubsection*{Combining all error estimates}
    Let $\mathbf{z}(t)$ and $\tilde{\mathbf{z}}(t)$ denote the solution of the neural ODE given initial conditions $\mathbf{z}
    $ and $\mathbf{z}+\delta_z$, respectively. Then \cite[Theorem 2.8]{teschl2012ordinary} yields $\norm{\mathcal{F}_t(\mathbf{z}) - \mathcal{F}_t(\mathbf{z}+\delta_z)}_2 \leq \norm{\delta_z}_2 e^{L_v t}$. Utilizing this stability estimate, we obtain,
    \begin{equation*} 
        \begin{aligned}
            \norm{ \mathcal{K}^\dagger(u_0, t) - \widehat{\mathcal{K}}^\dagger(U_0, t)}_\U & = \norm{D(\mathcal{F}_t(E(u_0))) - D(\widehat{\mathcal{F}}_t(\widehat{E}(U_0); \bm{t})}_\U \\
            & \leq L_D \norm{\mathcal{F}_t(E(u_0)) - \widehat{\mathcal{F}}_t(\widehat{E}(U_0); \bm{t})}_2 \\
            & \leq L_D \left(\norm{\mathcal{F}_t(E(u_0)) - \mathcal{F}_t(\widehat{E}(U_0))}_2 \right. \\
            & \left.  + \norm{\mathcal{F}_t(\widehat{E}(U_0)) - \widehat{\mathcal{F}}_t(\widehat{E}(U_0); \bm{t})}_2\right) \\
            & \leq L_D \left(  e^{L_v t} \norm{E(u_0) - \widehat{E}(U_0)}_2 + \mathcal{O}\left(\delta_t^{\min(4, q)}\right)\right) \\
            & \leq L_D \left(  e^{L_v t}L_{\mathcal{E}_\varphi} \norm{\mathrm{M}_\phi}_2 \norm{a^\dagger(u_0) - a(U_0)}_2  + \mathcal{O}\left(\delta_t^{\min(4, q)}\right)  \right), 
        \end{aligned}
    \end{equation*}
    where the second-to-last inequality follows from the stability estimate and Theorem \ref{thm:error_estimate_ode_discretization_p=2}, and the last inequality follows from the error estimate between $E$ and $\widehat{E}$. The full inequality now follows from Theorem \ref{lemma:general_coeff_error_estimate}.
\end{proof}
\begin{remark}
    Theorem \ref{thm:discretization_error_estimate_RONOM} assumes all involved neural networks are Lipschitz. This is reasonable, since common activation functions are Lipschitz, and \ref{SM:lipschitz_cont_of_NNs_and_RONOM} shows that feedforward neural networks built from such activations are Lipschitz. The same appendix also demonstrates that the full RONOM architecture is therefore Lipschitz.

    The Lipschitz behavior of the learned RONOM operator reflects the regularity of the underlying system. For instance, systems with large Lipschitz constants require networks with similarly large Lipschitz behavior, which makes training and robustness more challenging. When the target system is not Lipschitz, this appears to be at odds with the inherently Lipschitz nature of neural networks. In practice, however, training on finite datasets restricts attention to compact input domains on which the system may exhibit Lipschitz behavior. Overall, Lipschitz regularity provides useful insight into discretization robustness: larger constants amplify perturbations, helping to explain the observed variability in robustness across learned systems.
\end{remark}

\begin{remark}
The error estimate for the encoder depends on three main factors: the conditioning of the matrix $\Phi + \lambda \mathrm{L}$, and the approximation of the inner products in $\braket{u_0}{\bm{\phi}}_{L_2(\Omega)}$ and in $\Phi$. As shown in Section \ref{sec:error_estimates_integral_discretization}, these inner product approximation errors are influenced by domain approximation and numerical quadrature. Error estimates similar to those of our encoder also arise in FEM, where variational crimes account for errors due to, for instance, domain approximation, inexact matching of the boundary conditions, and numerical quadrature \cite{strang2008FEManalysis, brenner2008mathematical}.

Beyond the effects of domain approximation and numerical quadrature, the conditioning of $\Phi + \lambda \mathrm{L}$ and the FEM assembly matrix plays a critical role. In FEM, refining the mesh typically deteriorates the conditioning of the assembly matrix, which in turn can amplify quadrature errors when computing the right-hand side of the system. While FEM analysis focuses on how the condition number evolves with increasing mesh resolution, our approach employs a fixed global basis, rendering $\Phi$ independent of the mesh. Such construction requires only a single condition number analysis, irrespective of the discretization of the input function. 
    
\end{remark}

\section{Numerical experiments}
This section compares RONOM, FNO, CNO, and DeepONet based on the three properties in Figure~\ref{fig:different_tasks}, namely input generalization, super-resolution, and discretization robustness. The comparisons are demonstrated through three time-dependent PDE examples: the one-dimensional Burgers’ equation, the two-dimensional wave equation, and the two-dimensional incompressible Navier-Stokes equation in vorticity form.

To address time-dependency with the FNO and CNO, the time variable is added as a constant input feature at each spatial grid point. The trunk network takes both space and time as input for DeepONet. Since the data introduced in the next section is given on grids, RONOM’s encoder combines convolutional neural networks (CNNs), a grid-based sampling, and regularized projections. These projections use Gaussian basis functions centered at grid points and incorporate Sobolev norms, orthonormality penalties, and boundary constraints. The latent representation matches the spatial resolution, augmented with four channels; for example, it is $33 \times 4$ for Burgers’ equation and $33 \times 33 \times 4$ for the wave equation and Navier-Stokes equation. The explicit 4th-order Runge–Kutta method with fixed step is used for the time-stepping of the Neural ODE. The decoder maps latent codes back to the grid using a CNN and applies a Gaussian kernel for optimal recovery. More details on RONOM's design choices can be found in \ref{SM:ronom_arch_details}. For more details on the number of learnable parameters and computational costs of each model, we refer to Appendix \ref{app:model_size_and_computational_costs}.

\subsection{Data generation}
\subsubsection{Burgers' equation}
Consider a one-dimensional viscous Burgers' equation with periodic boundary conditions,
\begin{equation*}
\begin{cases}
    \frac{\partial u}{\partial t} + u\frac{\partial u}{\partial x}  = \nu \frac{\partial^2 u}{\partial x^2}, & (x, t) \in (0, 1) \times (0, 1]\\
    u(0,t) = u(1,t), & t \in (0, 1] \\
    \frac{\partial u}{\partial t}(0,t) =  \frac{\partial u}{\partial t}(1,t),  & t \in (0, 1]\\
    u(x,0) = u_0(x),  & x \in (0, 1)
\end{cases}
\end{equation*}
where $\nu = 0.01$ is the viscosity coefficient. The initial condition $u_0$ is generated from a Gaussian random field with mean zero, and as the covariance kernel, the operator $\sqrt{2}(250^2)(-\Delta + 25^2 I)^{-2}$ is considered with periodic boundary conditions. A total of 1000 solution trajectories are generated using Chebfun \cite{driscoll2014chebfun} on a spatial grid of size $1025$, with $101$ time points uniformly spaced between $0$ and $1$. To prepare the training data, the spatial and temporal grids are subsampled by factors of $32$ and $10$, respectively. This yields a spatial resolution of $33$, with time instances $[0.0, 0.1, \ldots, 1.0]$ in training. Among $1000$ solutions, $800$ of them are used for training and validation, while the remaining $200$ are for testing.

\subsubsection{Wave equation}
Consider a two-dimensional wave equation with wave propagation speed $c=0.3$, Dirichlet boundary conditions, initial condition $f$, and zero initial velocity:
\begin{equation*}
\begin{dcases}
\frac{\partial^2}{\partial t^2} u = c^2 \Delta u,  & (\bm{x},t) \in (0, 1)^2 \times (0,1] \\
  u(\bm{x},t) = 0, & (\bm{x}, t) \in \partial [0, 1]^2 \times [0, 1]\\
  u(\bm{x}, 0) = f(\bm{x}), & \bm{x} \in [0, 1]^2 \\
  u_t(\bm{x},0) = 0, & \bm{x} \in [0,1]^2,
\end{dcases} 
\end{equation*}
where $f(\bm{x}) = \sum_{i, j=1}^K a_{ij} (\pi^2 (i^2 + j^2))^{-r/2} \sin(\pi i x_1) \sin(\pi j x_2)$ denotes initial condition and the analytical solution of the PDE can be written as $u(\bm{x},t) = \sum_{i, j=1}^K a_{ij} (\pi^2 (i^2 + j^2))^{-r/2} \sin(\pi i x_1) \sin(\pi j x_2) \mathrm{cos}(c\pi t \sqrt{i^2 + j^2}),$
where $K=24$, $r=3$, $a_{11}=0$, and $a_{ij} \sim \mathcal{N}(0, 125^2)$. In this example, $1400$ solutions are generated based on a spatial grid of size $129 \times 129$, with $101$ time points uniformly spaced between $0$ and $1$. For training, the spatial grid is subsampled by a factor of 4 in each dimension, and the temporal grid is subsampled by a factor of 10. This yields a training resolution of $33 \times 33$, with time instances $[0.0, 0.1, \ldots, 1.0]$. Among the entire dataset, $1200$ of them are used for training and validation, while the remaining $200$ is used for testing.

\subsubsection{Incompressible Navier-Stokes equation}
Consider a two-dimensional incompressible Navier–Stokes equation in vorticity form with viscosity coefficient $\nu = 2 \times 10^{-4}$ and initial condition $w_0$,
\begin{equation*}
\begin{dcases}
\frac{\partial}{\partial t} w + \braket{\bm{u}}{\nabla w} = \nu \Delta w,  & (\bm{x},t) \in (0, 1)^2 \times (0,50], \\
 \nabla \cdot \bm{u}(\bm{x}, t) = 0, & (\bm{x},t) \in (0, 1)^2 \times (0,50], \\
  w(\bm{x}, 0) = w_0(\bm{x}), & \bm{x} \in [0, 1]^2 ,
\end{dcases} 
\end{equation*}
where periodic boundary conditions are imposed. The velocity field $\bm{u}$ is recovered from the vorticity through the stream function formulation.

The initial condition $w_0$ is generated from a Gaussian random field with mean zero and covariance operator $4^{7/2}(-\Delta + 16 I)^{-3.5}$ with periodic boundary conditions. Using $3400$ of such random initial conditions, we generate a total of $3400$ solution trajectories following the procedure described in the original FNO paper \cite{li2020fourier}. In particular, the method for solving the equations uses the stream function formulation, employs a pseudo-spectral method in space, and advances the solution in time using Crank-Nicolson updates with a time step of $10^{-3}$.

The solutions are computed on a spatial grid of size $129$ and stored at $101$ uniformly spaced time instances between $0$ and $50$. To construct the training data, the spatial and temporal grids are subsampled by factors of $4$ and $10$, respectively. This results in a spatial resolution of $33 \times 33$ and time instances $[0, 5, \ldots, 50]$ for training. Among the $3400$ trajectories, $3200$ are used for training and validation, while the remaining $200$ are for testing.

\begin{table}[b!]
    \caption{\textbf{\textit{Evaluating input generalization and super-resolution.}} We calculate the RMSE of the FNO, CNO, DeepONet, and RONOM predictions on the test dataset. The average values and standard deviations (between brackets) are of the order $10^{-3}$, and the best values are highlighted in bold. As FNO requires the input and output spatial grids to match, it can not be used for the spatial super-resolution experiment.}
    \begin{center}
    \captionsetup[subtable]{aboveskip=2pt, belowskip=-4pt}
    \begin{subtable}[b]{\textwidth}
        \caption{Burgers' Equation}
        \centering
        \begin{tabular}{|c||cc||cc||cc|}
        \hline
        \multirow{2}{*}{\textbf{Models}} & \multicolumn{2}{c||}{\textbf{Input}} & \multicolumn{4}{c|}{\textbf{Super-resolution}} \\
        \multicolumn{1}{|c||}{} & \multicolumn{2}{c||}{\textbf{generalization}} & \multicolumn{2}{c}{\textit{Time}}& \multicolumn{2}{c|}{\textit{Space + Time}}\\
        \cline{2-7}
        \hline \hline
        \textit{FNO} & \textbf{0.29} & (\textbf{0.51})  &\textbf{0.91} & (\textbf{2.76}) & $\times$ & $\times$\\
        \textit{CNO} & 1.12 & (1.93) &  1.63 & (5.75) & 9.38 & (28.56) \\
        \textit{DeepONet} & 4.91 & (5.13) &  5.09 & (9.22) & 5.38 & (12.44) \\
        \textit{RONOM} & 1.34 & (1.62)  & 1.72 & (4.92) & \textbf{1.73} &  (\textbf{5.23}) \\
        \hline
        \end{tabular}
    \end{subtable}
    \break
    \captionsetup[subtable]{aboveskip=2pt, belowskip=4pt}
    \begin{subtable}[b]{\textwidth}
        \caption{Wave Equation}
        \centering
        \begin{tabular}{|c||cc||cc||cc|}
        \hline
        \multirow{2}{*}{\textbf{Models}} & \multicolumn{2}{c||}{\textbf{Input}} & \multicolumn{4}{c|}{\textbf{Super-resolution}} \\
        \multicolumn{1}{|c||}{} & \multicolumn{2}{c||}{\textbf{generalization}} & \multicolumn{2}{c}{\textit{Time}}& \multicolumn{2}{c|}{\textit{Space + Time}}\\
        \cline{2-7}
        \hline \hline
        \textit{FNO} & \textbf{0.12} & (\textbf{2.79})  &\textbf{0.12} & (\textbf{2.98}) & $\times$ & $\times$\\
        \textit{CNO} & 0.17 & (2.80) &  0.59 & (7.44) & 5.75 & (24.41) \\
        \textit{DeepONet} & 5.71 & (9.50) &  5.51 & (9.30) & 5.57 & (9.88) \\
        \textit{RONOM} & 0.14 & (1.43)  & 0.33 & (5.00) & \textbf{0.37} &  (\textbf{5.18}) \\
        \hline
        \end{tabular}
    \end{subtable}
    \break
    \captionsetup[subtable]{aboveskip=2pt, belowskip=4pt}
    \begin{subtable}[b]{\textwidth}
        \caption{Incompressible Navier-Stokes Equation}
        \centering
        \begin{tabular}{|c||cc||cc||cc|}
        \hline
        \multirow{2}{*}{\textbf{Models}} & \multicolumn{2}{c||}{\textbf{Input}} & \multicolumn{4}{c|}{\textbf{Super-resolution}} \\
        \multicolumn{1}{|c||}{} & \multicolumn{2}{c||}{\textbf{generalization}} & \multicolumn{2}{c}{\textit{Time}}& \multicolumn{2}{c|}{\textit{Space + Time}}\\
        \cline{2-7}
        \hline \hline
        \textit{FNO} & \textbf{5.88} & (\textbf{8.55}) & \textbf{5.80} & (\textbf{10.28}) & $\times$ & $\times$\\
        \textit{CNO} & 7.72 & (14.98) &  7.76 & (17.40) &\textbf{8.71} & (\textbf{24.16}) \\
        \textit{DeepONet} & 15.08 & (45.50) &  15.20 & (46.20) & 15.22 & (46.22) \\
        \textit{RONOM} & 13.29 & (30.34)  & 13.35 & (30.90) & 13.34 &  (30.88) \\
        \hline
        \end{tabular}
    \end{subtable}
\end{center}
\label{tab:eval_metric_values}
\end{table}

\subsection{Results on input generalization and super-resolution}
In this section, the results concerning input generalization and super-resolution are presented. To demonstrate those two properties, the input discretization used during evaluation is kept consistent with the training resolution.

For input generalization, models are evaluated by mapping inputs and querying outputs at the same resolution as used during training. In the super-resolution setting, the input remains at the training resolution, while the output is generated at a higher spatial resolution. Specifically, this means generating solutions on the full grid with $1025$ points for the Burgers' equation and $129 \times 129$ grid points for the wave and Navier-Stokes equations. To assess the super-resolution performance regarding time, outputs are evaluated at both the training and full time resolutions. Table~\ref{tab:eval_metric_values} summarizes the quantitative results based on the root mean square error (RMSE).




\subsubsection{Input generalization}
Table~\ref{tab:eval_metric_values} shows that the FNO method outperforms both CNO and RONOM on the Burgers' equation, while CNO and RONOM achieve similar RMSE. All three methods perform comparably on the wave equation. On the Navier-Stokes equation, the FNO again performs best, whereas RONOM performs worst among the three. DeepONet performs the worst across all scenarios.

\begin{figure}[b!]
    \centering
    \includegraphics[width=0.77\linewidth]{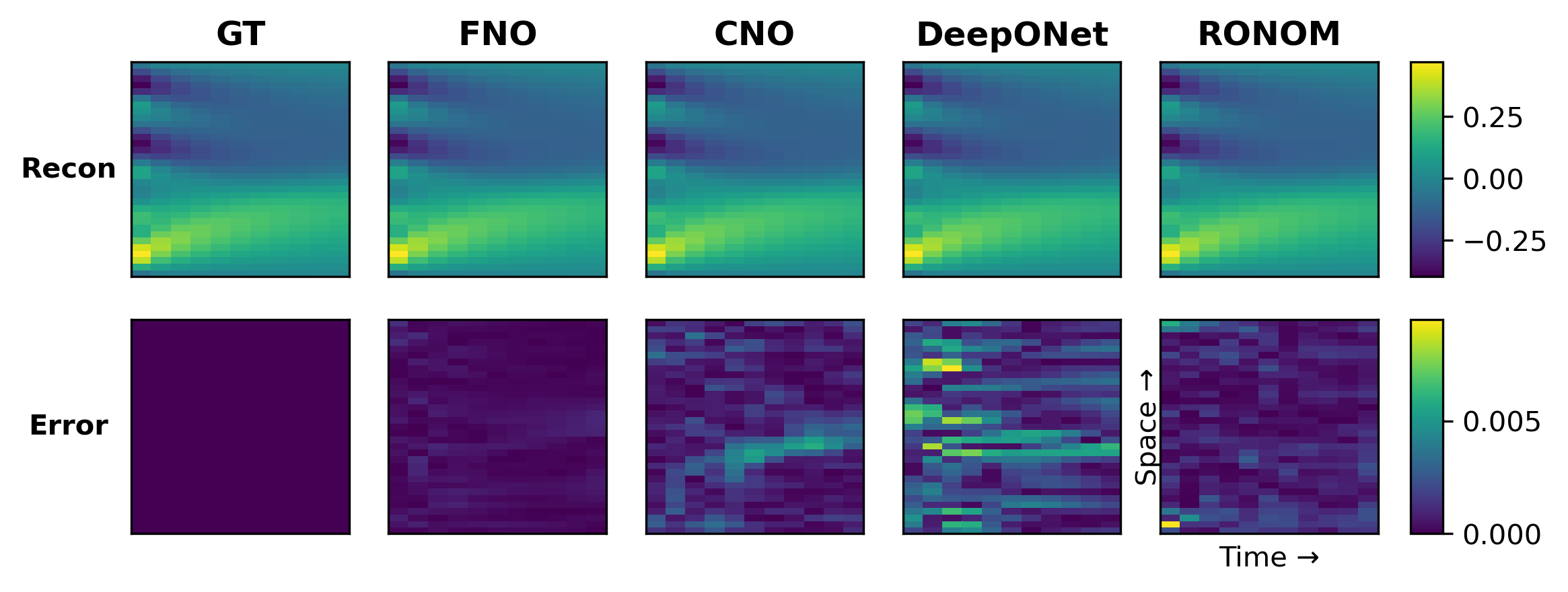}
    \caption{\textbf{\textit{Input generalization (Burgers' equation).}} Visualizations of FNO, CNO, DeepONet, and RONOM model predictions over time on the Burgers' equation. The ground truth (GT) solution and the corresponding discrepancy between GT and model prediction are also demonstrated for comparison.}
    \label{fig:input_generalization_burgers}
\end{figure}

\begin{figure}[t!]
    \centering
    \includegraphics[width=0.87\linewidth]{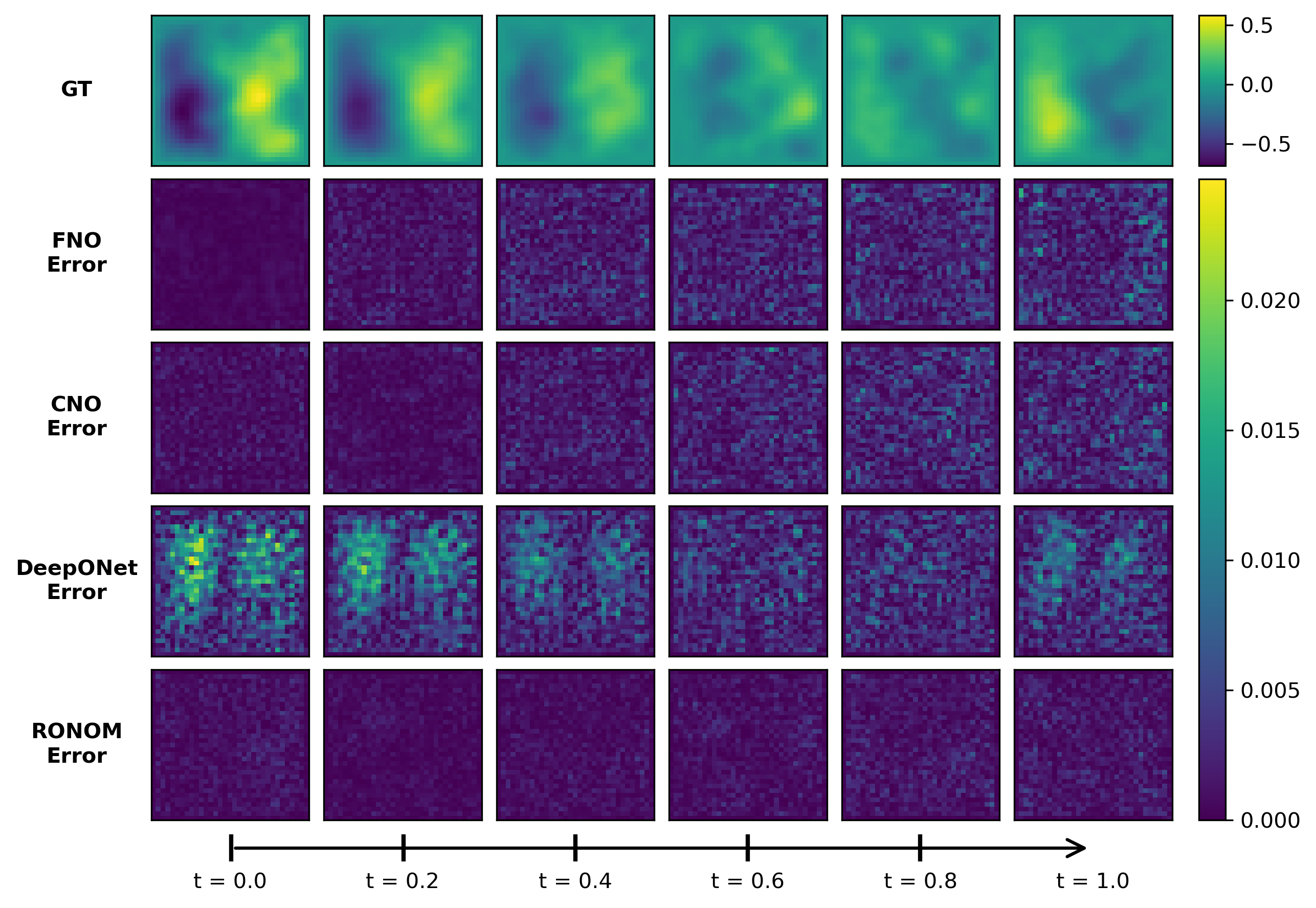}
    \caption{\textbf{\textit{Input generalization (wave equation).}} Visualization of prediction errors for FNO, CNO, DeepONet, and RONOM, shown alongside the ground truth (GT).}
    \label{fig:input_generalization_wave}
\end{figure}

\begin{figure}[t!]
    \centering
    \includegraphics[width=0.87\linewidth]{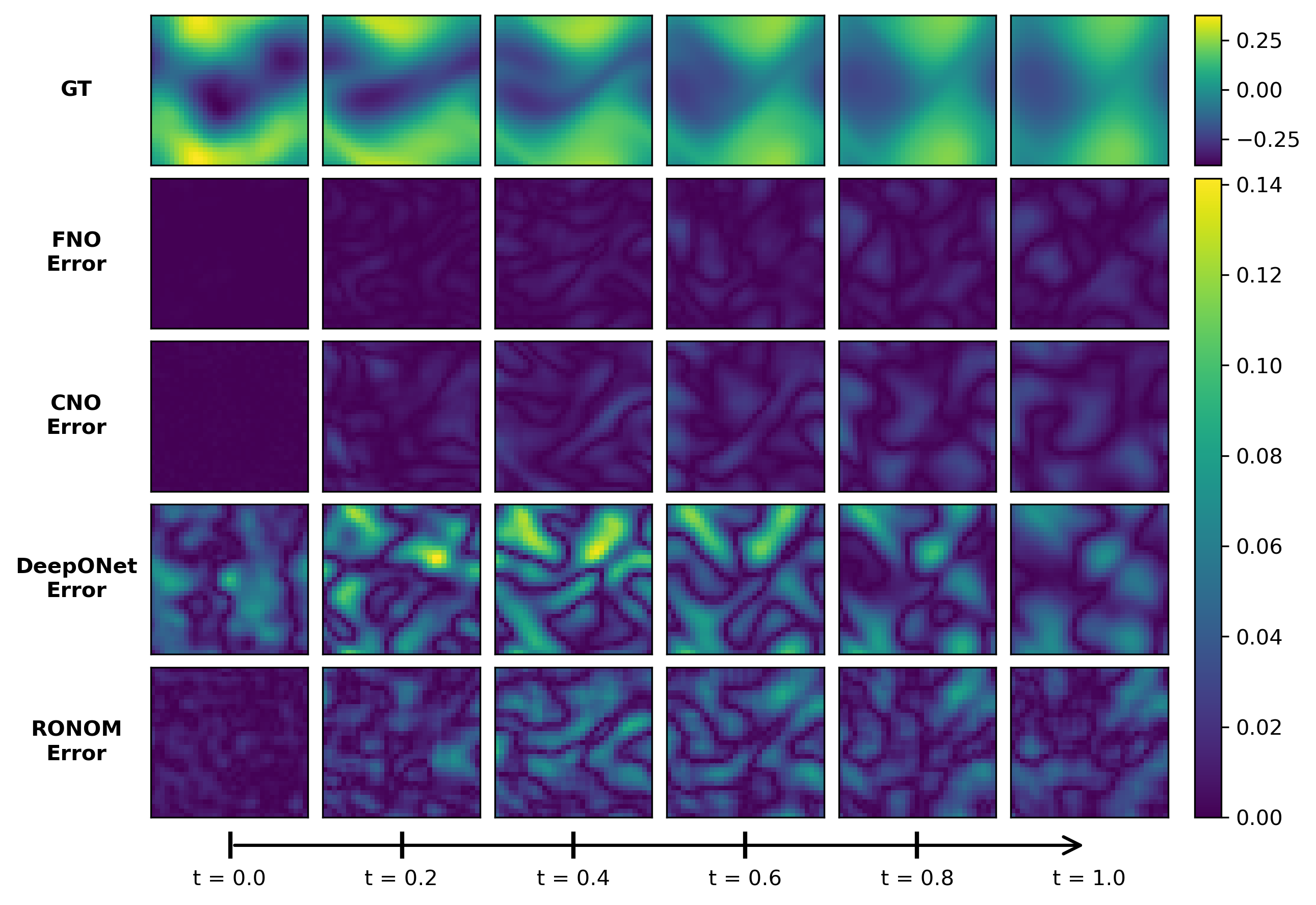}
    \caption{\textbf{\textit{Input generalization (Navier-Stokes).}} Visualization of prediction errors FNO, CNO, DeepONet, and RONOM, shown alongside the ground truth (GT).}
    \label{fig:input_generalization_NS}
\end{figure}

To further illustrate the differences among the methods, visualizations are shown in Figure~\ref{fig:input_generalization_burgers} for the Burgers' equation, in Figure~\ref{fig:input_generalization_wave} for the wave equation, and Figure~\ref{fig:input_generalization_NS} for Navier-Stokes. Figure~\ref{fig:input_generalization_burgers} shows that the FNO demonstrates much less reconstruction error, and it outperforms CNO, RONOM, and DeepONet on the Burgers' equation.  Moreover, both CNO and RONOM appear to struggle with accurately reconstructing the initial condition. This is also observed in the wave equation predictions presented in Figure~\ref{fig:input_generalization_wave}. The discrepancy in the initial condition reconstructions likely stems from the architectural differences since CNO and RONOM both rely on local convolutional layers, while the FNO allows copying the full initial condition via global Fourier layers. This highlights the potential benefit of explicitly enforcing the initial condition. Additionally, for the wave equation, the prediction error for FNO and CNO increases as time evolves. In contrast, the RONOM method maintains a relatively stable error over time. This suggests that the causal structure and inductive bias introduced by the neural ODE in RONOM help mitigate error propagation. 


Whereas RONOM performed similarly to FNO and CNO for the Burgers' and wave equations, Figure \ref{fig:input_generalization_NS} shows that FNO and CNO outperform DeepONet and RONOM on Navier–Stokes. Compared to Figures \ref{fig:input_generalization_burgers} and \ref{fig:input_generalization_wave}, the initial condition is also reconstructed less accurately. A likely reason for RONOM’s weaker performance is its neural ODE structure, as this is the main difference between RONOM and CNO: both are convolutional architectures, but CNO appends time as an input channel, whereas RONOM evolves a latent state via a neural ODE. Hence, while ROM-based neural ODEs introduced an inductive bias that helps mitigate error propagation in the wave equation, this bias might also hinder learning.

In conclusion, while FNO outperforms CNO and RONOM on the Burgers' equation, RONOM, which fundamentally is a classical vector-to-vector neural network, still demonstrates competitive input generalization on the Burgers' and wave equations. However, despite this competitive performance, the inductive bias of the ROM-based neural ODE can hinder learning and reduce performance. Hence, such natural inductive biases must be considered carefully in the context of input generalization.

\subsubsection{Super-resolution}

\begin{figure}[t!]
    \centering
    \includegraphics[width=0.87\linewidth]{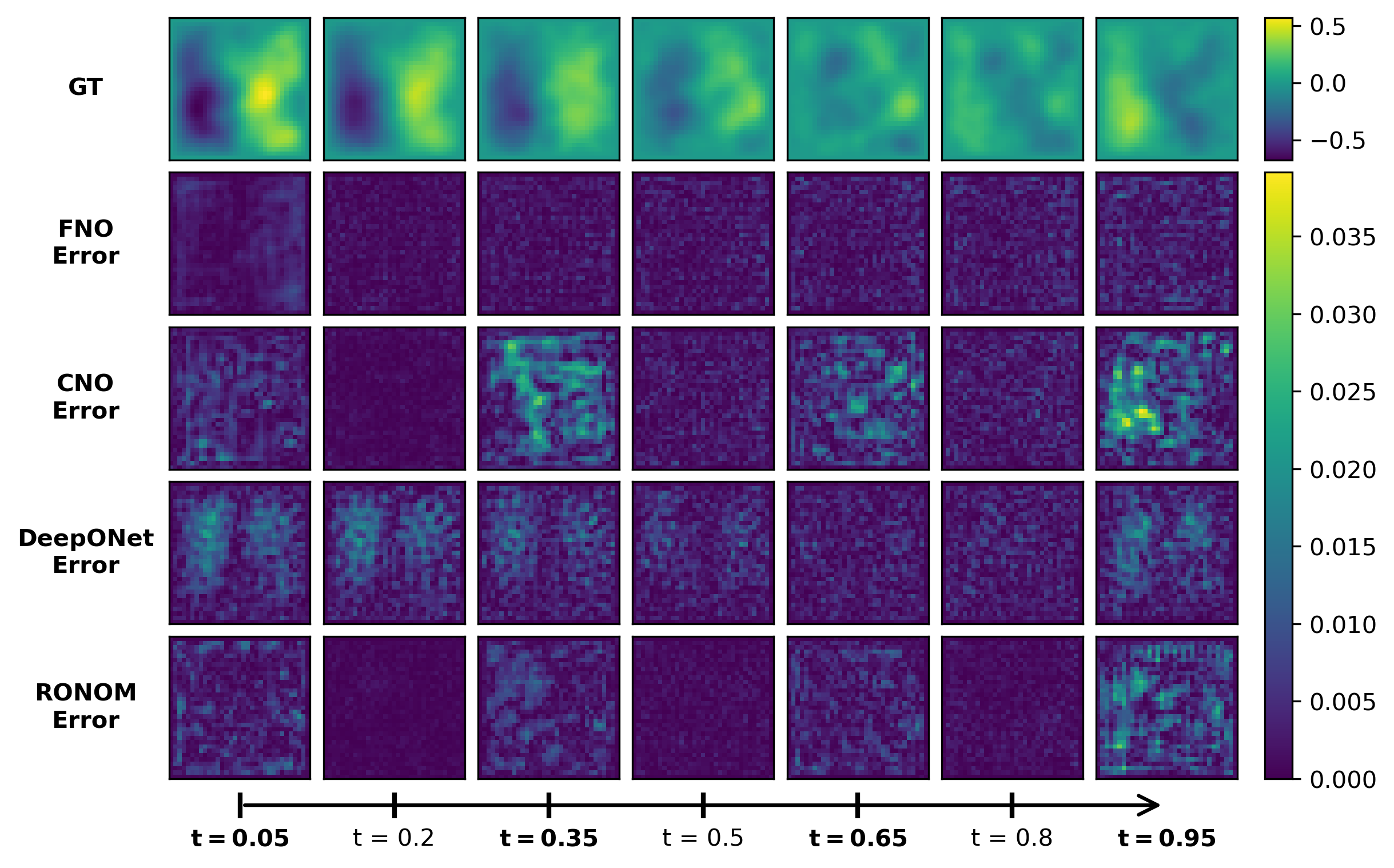}
    \caption{\textbf{\textit{Super-resolution (wave equation).}} Visualization of prediction errors for the FNO, CNO, DeepONet, and RONOM models, shown alongside the ground truth (GT). The initial condition, discretized on the training grid, is mapped to the output at training spatial resolution and full time resolution. Time instances highlighted in bold refer to the times that were not included during training.}
    \label{fig:superresolution_wave}
\end{figure}
The performance of RONOM, FNO, CNO, and DeepONet on super-resolution is also shown in Table~\ref{tab:eval_metric_values}. The results in the \textit{Time} column indicate that for Navier–Stokes, the reconstruction metrics change only marginally relative to the input generalization metrics. For the Burgers’ and wave equations, the metrics show more noticeable differences. Here, the FNO still performs best, while DeepONet performs worst.
CNO and RONOM demonstrate competitive performance with the FNO on the Burgers' equation. But the performance of both RONOM and CNO deteriorates more compared to that of the FNO on the wave equation.

Figure~\ref{fig:superresolution_wave} further illustrates such deterioration by demonstrating the error of the model predictions at unseen time instances. While FNO closely matches the intermediate solutions, both the CNO and RONOM exhibit significantly higher errors at these times. A key factor that leads to such behavior is the application of global Fourier layers versus local convolutional layers. The Fourier layers allow FNO to have global temporal interactions, while CNO and RONOM rely on local operations, which might be insufficient for generalizing in time. Another interesting observation from Table~\ref{tab:eval_metric_values} is that CNO appears to struggle with spatial upsampling, despite being specifically designed for this. This issue is further discussed in the next section, which focuses on robustness with respect to input discretization.

\subsection{Results on discretization robustness}
This section investigates the behaviors of the FNO, CNO, and RONOM models under varying discretizations of the input and output. Since DeepONet requires inputs remaining at the training resolution, it is excluded from this comparison. Furthermore, for a meaningful comparison of discretization effects, the methods must have similar input generalization. Otherwise, a method with poorer generalization may make discretization changes appear smaller, since they act on an already degraded reconstruction. This makes it harder to compare stability. Based on Table \ref{tab:eval_metric_values} and Figures \ref{fig:input_generalization_burgers}, \ref{fig:input_generalization_wave}, and \ref{fig:input_generalization_NS}, we therefore focus on the Burgers' and wave equations.

\begin{figure}[ht]
    \centering
    \begin{subfigure}[b]{0.48\textwidth}
        \centering
        \includegraphics[width=\textwidth]{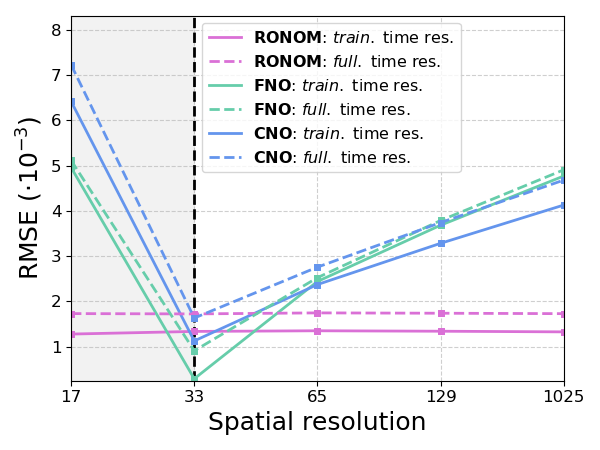}
        \caption{Burgers' equation}
        \label{fig:discr_robustness_graph_burgers}
    \end{subfigure}
    \hfill
    \begin{subfigure}[b]{0.48\textwidth}
        \centering
        \includegraphics[width=\textwidth]{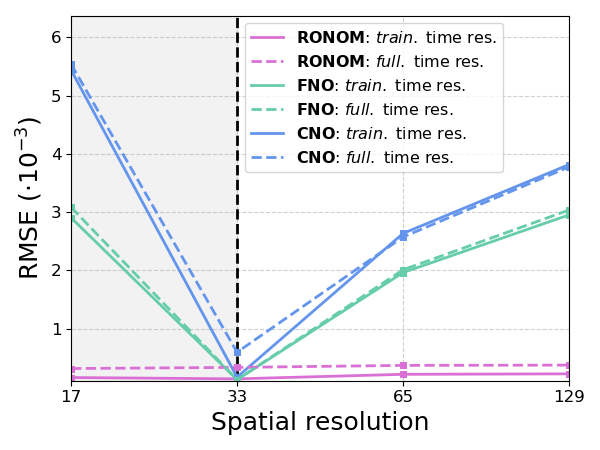}
        \caption{Wave equation}
        \label{fig:discr_robustness_graph_wave}
    \end{subfigure}
    \caption{\textbf{\textit{Discretization robustness performance}}. The quantitative performance of the FNO, CNO, and RONOM models is compared when discretizing the inputs and outputs at different spatial resolutions, with the dashed black line indicating the training spatial resolution. The results are shown both at the time discretization used during training and at the full time resolution available in the training data. RONOM is the only method robust to spatial resolution changes.}
\end{figure}

For the Burgers' equation, spatial subsampling by factors of 64, 32, 16, 8, and 1 is performed, where a factor of 32 is used for the training. For the wave equation, spatial subsampling is done by factors of 8, 4, 2, and 1, with a factor of 4 matching the training resolution. For the time variable, performance is evaluated both at the training resolution of 11 time points and at the full resolution of 101 time points. The results of this experiment for the Burgers' equation and wave equation are presented in Figure \ref{fig:discr_robustness_graph_burgers} and \ref{fig:discr_robustness_graph_wave}, respectively.

\begin{figure}[th]
    \centering
    \includegraphics[width=0.74\linewidth]{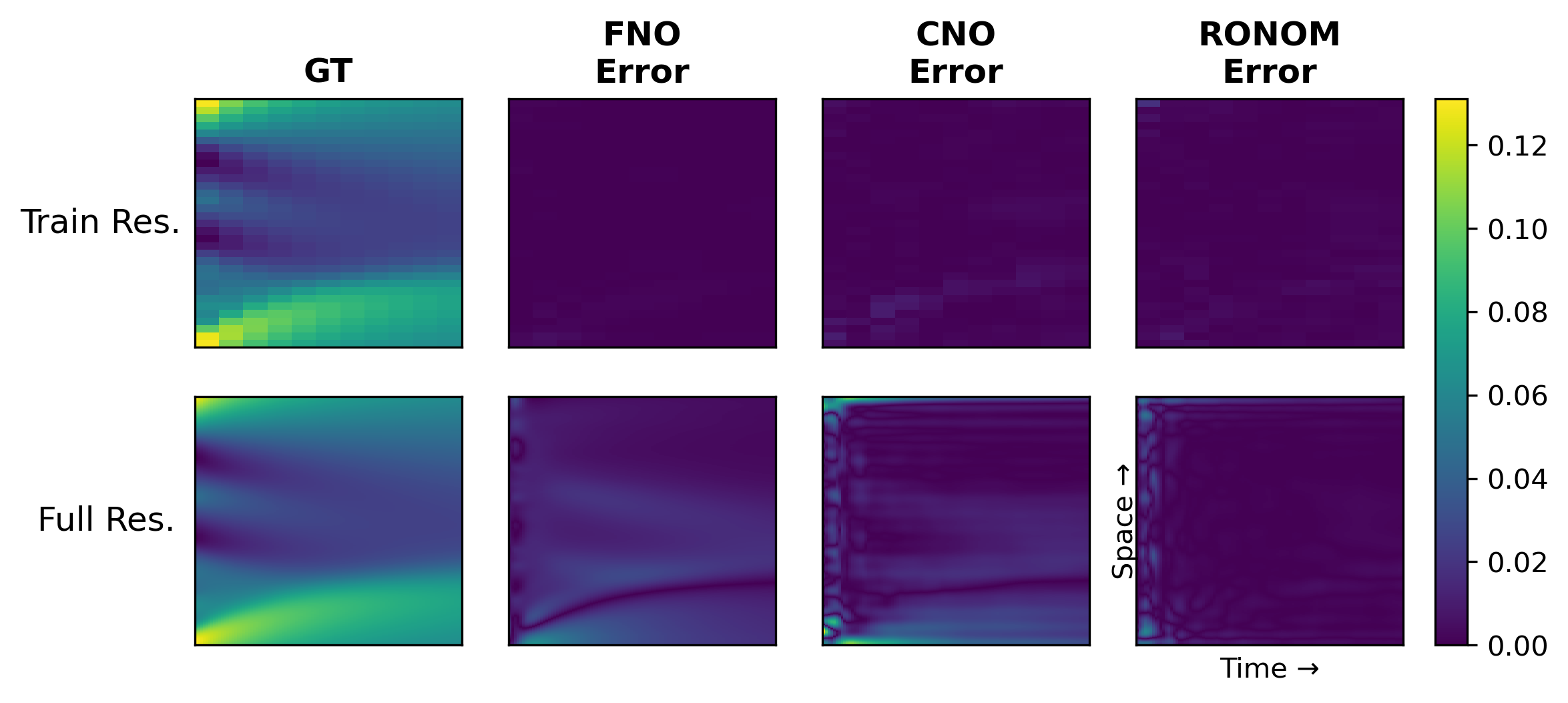}
    \caption{\textbf{\textit{Discretization effects (Burgers' equation).}} The first column presents the ground truth solution at the training resolution and the full resolution. The subsequent columns display pixel-wise reconstruction errors for FNO, CNO, and RONOM.}
    \label{fig:discr_robustness_burgers}
\end{figure}

Only the RONOM method is robust to changes in spatial input discretization. This indicates that simple convolutional architectures, as commonly used in deep learning-based ROMs, can achieve discretization robustness when equipped with appropriate lifting operations. In Raonic et al.~\cite{Raonic2023}, U-Net was found to be unsuitable for dealing with different input discretizations. It is mainly because spatial convolutions were defined on the training grid. When the input resolution increases, but the kernel size remains fixed, the convolution spans a smaller region in physical space. As a result, it effectively operates on smaller patches than it did during training, breaking consistency with the infinite-dimensional interpretation of the input function. By first lifting the input to a function and then querying this function at the training resolution, this inconsistency can be avoided, and it remains discretization robust.

\begin{figure}[th]
    \centering
    \includegraphics[width=0.985\linewidth]{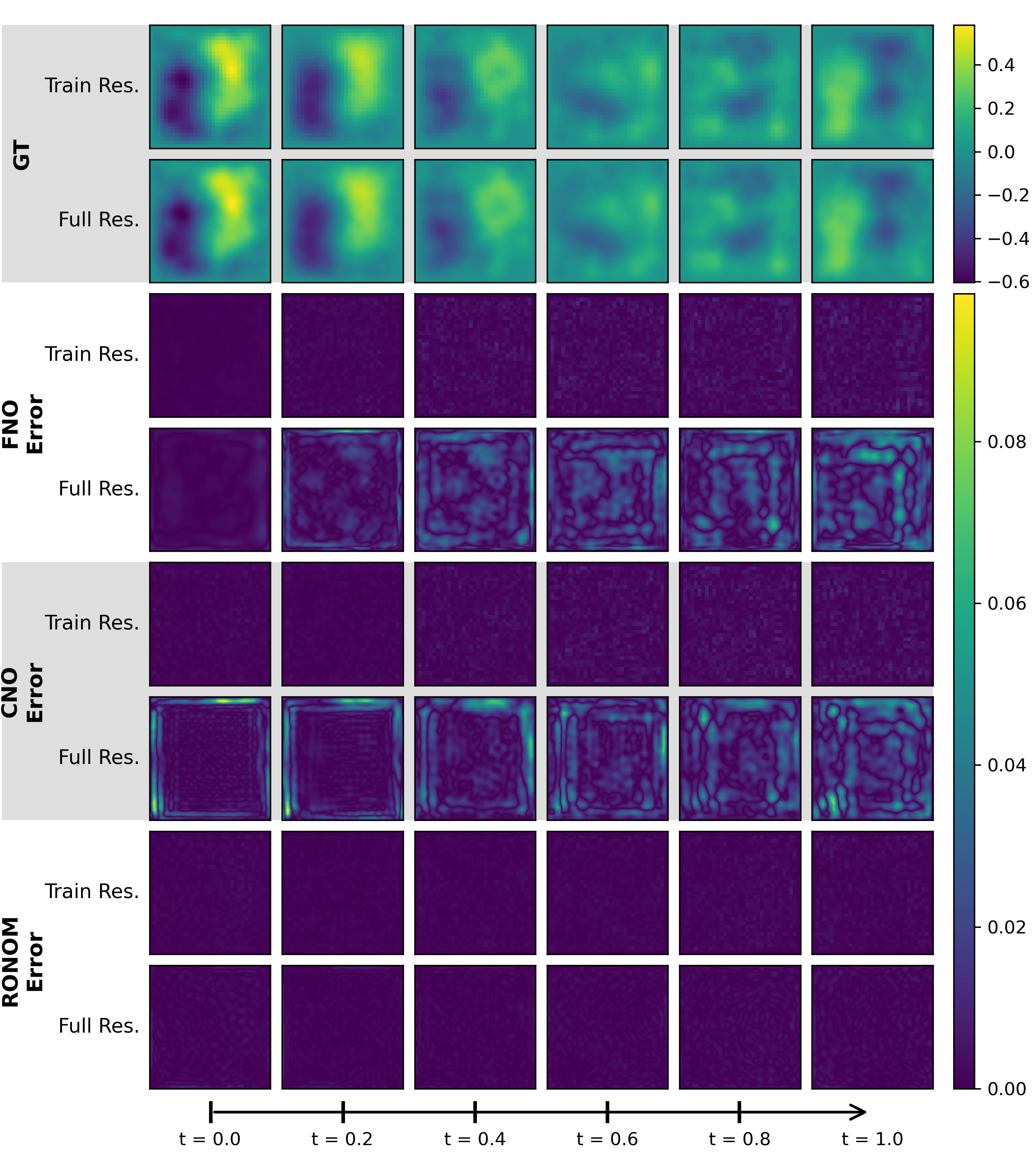}
    \caption{\textbf{\textit{Discretization effects (wave equation).}} The top row shows a ground truth example at the training resolution and the full data resolution. Below are the pixel-wise errors for the FNO, CNO, and RONOM methods. }
    \label{fig:discr_robustness_wave}
\end{figure}

To better understand why RONOM remains robust, while FNO and CNO do not, Figures~\ref{fig:discr_robustness_burgers} and~\ref{fig:discr_robustness_wave} display predictions for FNO, CNO, and RONOM on the Burgers' and wave equations, respectively. These figures confirm that, while the FNO achieves strong performance at the training resolution, it fails to generalize to the full spatial resolution. The CNO also exhibits poor performance at this higher resolution, primarily due to an inaccurate reconstruction of the initial condition.

In the CNO, inputs at resolutions different from the training resolution are resampled to the training resolution before processing. The resulting outputs are also resampled to match the desired output resolution. According to the Nyquist sampling theorem, a specific grid resolution is equivalent to a function in some space of bandlimited functions. Therefore, varying resolutions require mapping between different spaces of bandlimited functions. Specifically, consider a function $f$ whose Fourier transform satisfies $\operatorname{supp}(\hat{f}) \subseteq [-\omega, \omega]^d$. Upsampling to a space with a broader Fourier support, i.e. $[-\underline{\omega}, \underline{\omega}]^d$ with $\underline{\omega} \geq \omega$, corresponds to the identity operation as no new frequency components are introduced. When downsampling to a function with Fourier support in $[-\underline{\omega}, \underline{\omega}]^d$ where $\underline{\omega} < \omega$, CNO uses $\tilde{f}(x) = \left(\underline{\omega} / {\omega}\right)^d\int_\Omega h_{\underline{\omega}}(x-y)f(y) \dint y$ where $h_{\underline{\omega}}= \prod_{i=1}^d\sinc(2\underline{\omega} x_i)$ is the interpolation sinc filter with Fourier transform $\hat{h}_{\underline{\omega}} = \mathbbm{1}_{[-\underline{\omega}, \underline{\omega}]^d}$. Since convolution in the spatial domain corresponds to multiplication in the Fourier domain, this operation filters out the frequency components outside the new bandlimit. In practice, the CNO performs both upsampling and downsampling in Fourier space. Downsampling is achieved by truncating high-frequency components, while upsampling involves zero-padding the Fourier coefficients. This procedure uses the Fast Fourier Transform and only works properly when the source and target grid sizes are integer multiples of one another. When this condition is not met, the original Fourier coefficients correspond to incorrect (shifted) spatial frequencies in the new truncated or zero-padded representation. As a result, a different function is represented, producing errors compared to the original function. This makes the CNO sensitive to the specific input discretization. In contrast, our encoder uses a projection-based approach for up- and downsampling that is independent of grid structures. It supports arbitrary input locations, as long as the approximations of the inner products in Theorem~\ref{lemma:general_coeff_error_estimate} satisfy suitable error bounds (as discussed in Section~\ref{sec:error_estimates_integral_discretization}). This explains the robustness of our method compared to CNO and also accounts for the failure of the CNO on the spatial super-resolution experiment reported in Table~\ref{tab:eval_metric_values}. 

\subsection{Influence of the ODE solver on RONOM's performance}\label{sec:influence_ode_solver}
The previous section showed that RONOM is more robust to discretization than FNO and CNO. RONOM's results were obtained using the same ODE solver during training and inference. This raises the question of how robust RONOM is to changes in the time discretization within the ODE solver and to the choice of solver itself.

In this section, we investigate this question by fixing the ODE solver configuration during training and varying it at inference time. For both the Burgers' and wave equations, we use a fourth-order Runge-Kutta method with a time step of $0.1$ during training. At inference time, we evaluate the same RK4 solver and also consider it with a smaller time step of $0.001$. Furthermore, we consider the Euler method with time steps of $0.1$ and $0.001$, and the fifth-order adaptive Dormand Prince method (Dopri5). The corresponding results are reported in Table \ref{tab:dependence_RONOM_on_ODE_solver}.
\begin{table}[t]
    \centering
    \caption{\textbf{\textit{Influence of the ODE solver on RONOM's performance.}} We calculate RMSE values on the test dataset while varying the numerical ODE solver and its stepsize within RONOM. The average values and standard deviations (between brackets) are of the order $10^{-3}$, and the solver configuration used during training is highlighted in bold.}
    \begin{center}
    \captionsetup[subtable]{aboveskip=2pt, belowskip=-4pt}
    \begin{subtable}[b]{\textwidth}
        \caption{Burgers' Equation}
        \centering
        \begin{tabular}{|cc||cc||cc||cc|}
            \hline
            \multirow{2}{*}{\textbf{Solver}} & \multirow{2}{*}{\textbf{Stepsize}} & \multicolumn{2}{c||}{\textbf{Input}} & \multicolumn{4}{c|}{\textbf{Super-resolution}} \\
            & & \multicolumn{2}{c||}{\textbf{generalization}} & \multicolumn{2}{c}{\textit{Time}}& \multicolumn{2}{c|}{\textit{Space + Time}}\\
            \cline{2-7}
            \hline \hline
            \textbf{\textit{RK4}} & \textbf{0.1} &\textbf{1.34} & \textbf{(1.62)}  & \textbf{1.72} & \textbf{(4.92)} & \textbf{1.73} &  \textbf{(5.23)} \\
            \textit{RK4} & 0.001 & 1.34 & (1.84) & 1.69 & (4.71) & 1.69 & (5.01)\\
            \textit{Euler} & 0.1 & 1.11 & (10.79) & 1.46 & (10.86) & 1.50 & (10.99)\\
            \textit{Euler} & 0.001 & 1.34 & (1.84) & 1.69 & (4.70) & 1.69 & (5.00)\\
            \textit{Dopri5} & \textit{Adaptive} & 1.34 & (1.84) & 1.69 & (4.71) & 1.69 & (5.01)\\
            \hline
        \end{tabular}
    \end{subtable}
    \break
    \captionsetup[subtable]{aboveskip=2pt, belowskip=4pt}
    \begin{subtable}[b]{\textwidth}
        \caption{Wave Equation}
        \centering
        \begin{tabular}{|cc||cc||cc||cc|}
            \hline
            \multirow{2}{*}{\textbf{Solver}} & \multirow{2}{*}{\textbf{Stepsize}} & \multicolumn{2}{c||}{\textbf{Input}} & \multicolumn{4}{c|}{\textbf{Super-resolution}} \\
            & & \multicolumn{2}{c||}{\textbf{generalization}} & \multicolumn{2}{c}{\textit{Time}}& \multicolumn{2}{c|}{\textit{Space + Time}}\\
            \cline{2-7}
            \hline \hline
            \textbf{\textit{RK4}} & \textbf{0.1} & \textbf{0.14} & \textbf{(1.43)} & \textbf{0.33} & \textbf{(5.00)} & \textbf{0.37} &  \textbf{(5.18)} \\
            \textit{RK4} & 0.001 & 2.29 & (8.56) & 2.68 & (11.43) & 2.76 & (11.71)\\
            \textit{Euler} & 0.1 & 3.33 & (16.31) & 2.93 & (16.09) & 3.01 & (16.45)\\
            \textit{Euler} & 0.001 & 2.29 & (8.57) & 2.68 & (11.44) & 2.76 & (11.72)\\
            \textit{Dopri5} & \textit{Adaptive} & 2.29 & (8.56) & 2.68 & (11.37) & 2.76 & (11.71)\\
            \hline
        \end{tabular}
    \end{subtable}
\end{center}
    \label{tab:dependence_RONOM_on_ODE_solver}
\end{table}

For the Burgers' equation, Table \ref{tab:dependence_RONOM_on_ODE_solver} shows that our method is stable with respect to variations in the ODE solver and the step size. The most notable difference occurs when using the Euler method with the same step size as during training. In this case, the average RMSE is lower. However, because the solver is less accurate and therefore produces more variability across samples, the standard deviation is higher. The situation is different for the wave equation. While solvers with sufficiently small step sizes yield similar performance and therefore approximate the ODE comparably well, their performance is worse than that of the configuration used during training. 

These results indicate that the effect of the ODE solver configuration depends on the problem. For the wave equation, it matters, whereas for the Burgers' equation, it does not. There are several possible explanations. First, the reconstruction metrics for the wave equation are significantly lower compared to the Burgers' equation. Hence, even small deviations in the solution to the ODE can have a noticeable effect. Returning to Theorem \ref{thm:discretization_error_estimate_RONOM}, and assuming the same encoding, another explanation is that the vector field in the wave equation is more irregular. Consequently, the RK4 method with a step size of 0.1 deviates more from the underlying continuous-time dynamics. Another possible source of error is a larger Lipschitz constant of the decoder.

\section{Conclusion}
ROM and neural operators are both widely used to speed up the computation of solutions to time-dependent PDEs. While ROM methods rely on a fixed input discretization, neural operators are designed to generalize across varying discretizations. However, neural operator approaches do not explicitly quantify how well the discretized operators approximate their learned infinite-dimensional counterparts. In contrast, the ROM framework explicitly accounts for the discretization error of the associated ordinary differential equations. This work bridges the concepts from ROM and operator learning by introducing the reduced-order neural operator modeling (RONOM) framework. A general error estimate is derived to quantify the discrepancy between the learned infinite-dimensional operator and its approximation, explicitly accounting for errors arising from both the input function discretization and the neural ODE discretization. The error estimate guarantees discretization convergence of our model. Furthermore, RONOM is compared to FNO, CNO, and DeepONet on three numerical examples on three key aspects: input generalization, super-resolution, and robustness to discretization changes. The results show that RONOM, which uses standard vector-to-vector neural networks, can achieve competitive generalization and exhibits strong performance in spatial super-resolution and discretization robustness, where FNO and CNO underperform. In particular, the generalization results shed light on how ROM-based neural ODEs compare with alternative neural operator learning approaches for dealing with time. Moreover, our findings on temporal super-resolution provide insights for improving temporal accuracy in operator learning.

\section{Future work}
To further leverage the knowledge about theory and computational aspects of ROM, it could be valuable to further explore ROM's low-dimensional representation and the related Kolmogorov n-width. In particular, our methodology could benefit from incorporating structures into the learned latent space, like structure-preserving ROMs. These ROMs retain physical or geometric properties, such as energy conservation, symplecticity, or passivity in the reduced system to ensure stability and accuracy over long time horizons \cite{POLYUGA2010665,peng2016symplectic,buchfink2023symplectic}. Physics-informed techniques \cite{li2024physics,wang2021learning, dummer2025joint} may also be useful by explicitly incorporating the PDE into the loss function. Moreover, ROM sometimes explicitly enforces the initial condition, a feature not currently implemented in our approach. More broadly, integrating additional low-dimensional, structure-preserving, and physics-informed strategies within our RONOM offers a promising direction for future work.

Furthermore, the PDEs considered in this work are defined on square domains. However, the encoder and decoder in RONOM apply to arbitrary domains. It could therefore be valuable to investigate RONOM for PDEs defined on more complex or even varying geometries. More broadly, extending the RONOM framework to a wider range of PDE problems is a promising direction for future work.

Finally, compared to the fixed bases used in our approach, using bases that adapt to input discretization, such as those in FEM, could yield more accurate reconstructions of the input function. While this adaptability offers potential benefits, it also complicates the analysis of discretization errors: not only is the functional discretized, but the function space itself becomes variable. In future work, we plan to investigate adaptive FEM bases and aim to develop discretization error estimates by leveraging two sources of information. The first comes from error estimates associated with the discretization of the optimization functional. The second comes from FEM analyses that assess how variations in the optimization space influence the results.

\subsection*{Competing interests}
All authors declare that they have no conflicts of interest.

\subsection*{Data availability}
All the data and source codes to reproduce the results will be made available on GitHub at \url{https://github.com/SCdummer/RONOM.git}.

\appendix
\section{Proof of Theorem \ref{thm:integral_error_estimate}} \label{app:integral_error_estimate}
Before proving the theorem, we state and prove the following lemma
\begin{lemma}
    With $\lambda$ the Lebesgue measure, assume $\Omega = \bigcup_{k=1}^N \Omega_k$ with $\lambda(\Omega_i \cap \Omega_j) = 0$ for $i \neq j$ and $\lambda(\Omega_k) > 0$. Given a $u \in \TV(\Omega)$ and denoting the restriction to $\Omega_k$ as $u \mid_{\Omega_k} \colon \Omega_k \to \R$, the following holds,
    \begin{equation*}
        \TV(u) := \sup_{\substack{\phi \in C_c^\infty (\Omega)\\ \norm{\phi}_{L_\infty(\Omega)}\leq 1}} \int_\Omega u(\bm{x}) \nabla \cdot \phi(\bm{x}) \mathrm{d}\bm{x} \geq \sum_{k=1}^N \TV(u \mid_{\Omega_k}).
    \end{equation*} 
    \label{lemma:TV_inequality}
\end{lemma}
\begin{proof}
     As $\TV(u|_{\Omega_k}) \leq \TV(u) < \infty$ due to $\Omega \subseteq \Omega_k$, by the definition of the total variation, we have for every $\epsilon > 0$ a $\phi_k^\epsilon \in C_c^\infty(\Omega_k)$ such that $\int_{\Omega_k} u(\bm{x}) \nabla \cdot \phi_k^\epsilon(\bm{x}) \mathrm{d}\bm{x} \geq \TV(u\mid_{\Omega_k}) - \epsilon$. Define $\phi^\epsilon(\bm{x}):=\sum_{k=1}^N \phi_k^\epsilon(\bm{x})$ with $\phi_k^\epsilon(\bm{x})=0$ if $\bm{x} \in \Omega / \Omega_k$. Note that $\phi^\epsilon \in C_c^\infty(\Omega)$ and $\phi^\epsilon \mid_{\Omega_k} = \phi_k^\epsilon$ as $\lambda(\Omega_i \cap \Omega_j) = 0$ for $i \neq j$. Hence, we obtain:
    \begin{equation*}
        \begin{aligned}
            \TV(u) & \geq \int_\Omega u(\bm{x}) \nabla \cdot \phi^\epsilon(\bm{x}) \mathrm{d}\bm{x} = \sum_{k=1}^N \int_\Omega u(\bm{x}) \nabla \cdot \phi_k^\epsilon(\bm{x}) \mathrm{d}\bm{x} = \sum_{k=1}^N \int_{\Omega_k} u(\bm{x}) \nabla \cdot \phi_k^\epsilon(\bm{x}) \mathrm{d}\bm{x} \\
            & \geq \sum_{k=1}^N \TV(u \mid_{\Omega_k}) - \epsilon = \left(\sum_{k=1}^N \TV(u \mid_{\Omega_k})\right) - \epsilon N.
        \end{aligned}
    \end{equation*}
    Letting $\epsilon \rightarrow 0$ yields the desired inequality. 
\end{proof}


\begin{proof}[Proof Theorem \ref{thm:integral_error_estimate}]
    Note that: 
    \begin{equation}
        \begin{split}
        \int_\Omega f(\bm{x}) \dint \bm{x} - \sum_{k=1}^N \left( \sum_{i=1}^{m_k} w_{i,k} f(\bm{x}_{i,k}) \right) & = \left(\int_\Omega f(\bm{x}) \dint \bm{x} - \int_{\Omega^N} f(\bm{x}) \dint \bm{x}\right) + \\ 
        & \left(\int_{\Omega^N} f(\bm{x}) \dint \bm{x} -  \sum_{k=1}^N \left( \sum_{i=1}^{m_k} w_{i,k} f(\bm{x}_{i,k}) \right)\right).
        \end{split}
        \label{ineq:decomp_of_error}
    \end{equation}
    The first part of the error (domain approximation error) can be bounded as follows,
    \begin{align}
        \left|\int_\Omega f(\bm{x}) \dint \bm{x}  - \int_{\Omega^N} f(\bm{x}) \dint \bm{x} \right| & \leq \int_{\Omega \Delta \Omega^N} |f(\bm{x})|\dint \bm{x} \leq |\Omega \Delta \Omega^N| \norm{f}_{L_\infty(\Omega)},
        \label{ineq:domain_approx}
    \end{align}
    where the final equality follows from the fact that $\TV$ functions are essentially bounded and $C^{p+1}(\Omega)$ functions on a compact domain are bounded. In the second part of the error, the element-wise integral for $f \in \TV(\Omega)$ can be decomposed into two parts,
    \begin{align*}
            \int_{\Omega_k} f(\bm{x}) \dint \bm{x} & = \sum_{i=1}^{m_k} \widetilde{w}_{i,k} \int_{\Omega_k} f(\bm{x}) \dint \bm{x}  = \sum_{i=1}^{m_k} \widetilde{w}_{i,k} \int_{\Omega_k} f(\bm{x}_{i,k}) + \left( f(\bm{x}) - f(\bm{x}_{i,k}) \right) \dint \bm{x} \\
            & = |\Omega_k| \left(\sum_{i=1}^{m_k} \widetilde{w}_{i,k} f(\bm{x}_{i,k}) \right) + \int_{\Omega_k} \sum_{i=1}^{m_k} \widetilde{w}_{i,k}\left( f(\bm{x}) - f(\bm{x}_{i,k}) \right) \dint \bm{x}. 
    \end{align*}
    As $u \in \TV(\Omega)$, we know $u$ is essentially bounded. Hence, there exists $c_k \leq f(\bm{x}) \leq C_k$ for $\bm{x} \in \Omega_k$. We obtain that,
    \begin{align*}
            \left|\int_{\Omega_k} f(\bm{x}) \dint \bm{x} -   \left(\sum_{i=1}^{m_k} w_{i,k} f(\bm{x}_{i,k}) \right)\right| & = \left|\int_{\Omega_k} f(\bm{x}) \dint \bm{x} -  |\Omega_k| \left(\sum_{i=1}^{m_k} \widetilde{w}_{i,k} f(\bm{x}_{i,k}) \right)\right| \\
            & = \left|\sum_{i=1}^m \widetilde{w}_{i,k} \int_{\Omega_k} (f(\bm{x}) - f(\bm{x}_{i,k})) \dint \bm{x} \right| \\
            & \leq \sum_{i=1}^m \widetilde{w}_{i,k} \int_{\Omega_k} (C_k - c_k) \dint \bm{x} \\
            & = |\Omega_k| (C_k - c_k)  \leq |\Omega_k| \TV(f\mid_{\Omega_k}) ,
    \end{align*}
    where the first inequality uses $c_k \leq f(\bm{x})\leq C_k$ for $\bm{x} \in \Omega_k$ and the final inequality uses the definition of the total variation. Furthermore, from \cite{ern2004theory}, it follows that for $f \in C^{p+1}(\Omega)$, $\left| \int_{\Omega_k} f(\bm{x}) \dint \bm{x} - \sum_{i=1}^{m} w_{i,k} f(\bm{x}_{i,k}) \right| \leq c |\Omega_k| h_{\Omega_k}^{p+1} \sup_{\substack{\bm{x} \in \Omega_k \\ \sum_{j=1}^d \gamma_j = p+1}} \left|\partial^\gamma f(\bm{x}) \right|$. Utilizing this, the following error estimate on $\Omega^N$ is obtained, 
    \begin{align*}
            \left|\int_{\Omega^N} f(\bm{x}) \dint \bm{x} -  \sum_{k=1}^N \left( \sum_{i=1}^{m_k} w_{i,k} f(\bm{x}_{i,k}) \right)\right| & = \left|\sum_{k=1}^N \int_{\Omega_k} f(\bm{x}) \dint \bm{x} -  \left(\sum_{i=1}^{m_k} w_{i,k}  f(\bm{x}_{i,k}) \right)\right|\\
            & \leq \sum_{k=1}^N \left|\int_{\Omega_k} f(\bm{x}) \dint \bm{x} -  \left(\sum_{i=1}^{m_k} w_{i,k} f(\bm{x}_{i,k}) \right)\right| \\
            & \leq \begin{cases}
                c\sum\limits_{k=1}^N  h_{\Omega_k}^{p+1} |\Omega_k| \sup_{\substack{\bm{x} \in \Omega_k \\ \sum_{i=1}^d \gamma_i = p+1}} \left|\partial^\gamma f(\bm{x}) \right| \\ 
                \sum\limits_{k=1}^N |\Omega_k| \TV(f \mid_{\Omega_k}) 
            \end{cases} \\
            & \leq \begin{cases}
                c h^{p+1} |\Omega^N| \sup_{\substack{\bm{x} \in \Omega \\ \sum_{i=1}^d \gamma_i = p+1}} \left|\partial^\gamma f(\bm{x}) \right| \\
                h \TV(f),
            \end{cases}
    \end{align*}
    where the first case corresponds to $f \in C^{p+1}(\Omega)$ and the second to $f \in \TV(\Omega)$. The second inequality uses the error estimates on $\Omega_k$, and the last inequality follows from Lemma \ref{lemma:TV_inequality} and $h_{\Omega_k} \leq h$. By combining the above estimate with inequalities \eqref{ineq:decomp_of_error} and \eqref{ineq:domain_approx}, the desired error estimate is obtained. 
\end{proof}

\section{Model size and computational costs}\label{app:model_size_and_computational_costs}
This appendix provides additional details on the different models by examining their size and computational cost. The number of model parameters is reported in Table \ref{tab:number_of_model_parameters}, while the computational costs are summarized in Table \ref{tab:computational_costs}.
\begin{table}[t]
    \centering
    \caption{\textbf{\textit{Number of learnable model parameters}} for the models applied to the Burgers' and wave equations.}
    \begin{tabular}{|c||c||c||c|}
        \hline
        \textbf{Models} & \textbf{Burgers} & \textbf{Wave} \\
        \hline \hline
        \textit{FNO} & 806273 & 9653889 \\
        \textit{CNO} & 219281 & 655121  \\
        \textit{DeepONet} & 4253697 & 4794369  \\
        \textit{RONOM} & 414217 & 1734025  \\    
        \hline
    \end{tabular}
    \label{tab:number_of_model_parameters}
\end{table}

\begin{table}[t]
    \centering
    \caption{\textbf{\textit{Computational costs.}} We report the average time in seconds to process a single test instance of the Wave equation for different spatial and temporal resolutions. For RONOM, costs are reported for both the ODE solver used during training and the fifth-order adaptive Dormand–Prince (Dopri5) solver. The reported costs for RONOM exclude the LU decomposition for the discretized projection, as it is computed once per grid size and is reused.}
    \begin{tabular}{|c||c||c||c|}
        \hline
        \multirow{2}{*}{\textbf{Models}} & \textbf{Space: 33 x 33} & \textbf{Space: 33 x 33} & \textbf{Space: 129 x 129}  \\
         & \textit{Time: 10} & \textit{Time: 101} & \textit{Time: 101} \\
        \cline{2-4}
        \hline \hline
        \textit{FNO} & 0.003604 & 0.012967 & 0.332334\\
        \textit{CNO} & 0.004394 &  0.012506 & 0.024568  \\
        \textit{DeepONet} & 0.002608 & 0.029451 & 0.484781 \\
        \textit{RONOM (train)} & 0.014335 & 0.032622 & 0.047510 \\
        \textit{RONOM (Dopri5)} & 0.245796 & 0.256659 & 0.276529 \\
        \hline
    \end{tabular}
    \label{tab:computational_costs}
\end{table}

\subsection{Model size}
For the number of parameters presented in Table \ref{tab:number_of_model_parameters}, we focus on the Burgers' equation and the wave equation. We omit the Navier–Stokes case, since its model configuration coincides with that of the wave equation. One notable observation from Table \ref{tab:number_of_model_parameters} is that the number of parameters of the FNO is the most sensitive to the dimension of the spatial domain. The most important weights of the FNO parameterize a function in the frequency domain by specifying a fixed number of frequency modes in each spatial dimension. Since a sufficient number of modes is required, increasing the spatial dimension causes the number of weights to grow rapidly. This global weight structure contrasts with convolutional architectures, where kernels are typically small and local. The number of parameters, therefore, scales much more mildly. This difference between global and local convolutions can be observed in the number of parameters in RONOM and CNO, which both use local convolutions. 

\subsection{Computational costs}
For the analysis of computational cost, we focus on the wave equation. We exclude the Navier-Stokes equation because it uses the same model configuration and shares the same data shape as the wave equation, with 101 time points and a 129 by 129 spatial grid. The Burgers' equation is omitted since it is one-dimensional rather than two-dimensional. Specifically, it has 1025 spatial grid points and 101 time points, whereas the wave equation is defined on a 129 by 129 grid with 101 time points. As a result, even when subsampled, the wave equation remains significantly more computationally demanding due to its two-dimensional structure and larger total number of grid points. For these reasons, it suffices to restrict the computational cost comparison to the wave equation.

Table \ref{tab:computational_costs} shows that CNO and RONOM have relatively stable computational costs as the temporal and spatial resolutions increase, whereas FNO is strongly affected by increases in spatial resolution. This difference primarily arises from the use of local convolutions in CNO and RONOM, compared to the global Fourier-based convolutions in FNO, whose costs grow significantly with the input size. DeepONet is even more sensitive, as its cost scales directly with the number of evaluation points. In particular, increasing the temporal resolution by a factor of ten leads to ten times more evaluations. Overall, RONOM is the most computationally expensive method due to the ODE solve, although this cost depends on the chosen solver and can be tuned accordingly.

This computational cost analysis should be interpreted with care. For RONOM, the reported costs do not include the computation of the LU decomposition used to solve the discretized projection problem. When working on a fixed grid across many inputs, this decomposition can be reused and does not need to be recomputed. However, changing the grid resolution requires recomputing the LU decomposition, which can be costly for large grids and should be taken into account when interpreting these results.

\bibliographystyle{siamplain}
\bibliography{references}
\end{document}


\maketitle

\section{Additional theorems and proofs}

\subsection{Extension of Theorem \ref{thm:error_estimate_ode_discretization_p=2}} \label{SM:p_unequal_to_2_case_error_estimate_NODE}
This section considers general Hermite interpolation instead of only cubic Hermite interpolation. The Hermite interpolation uses $\widehat{\mathbf{z}}(t_i)$ and the exact higher-order time derivatives computed from the ODE vector field. As discussed in the proof of Theorem \ref{thm:error_estimate_ode_discretization_p=2}, we have $\frac{\mathrm{d}^{k}}{\mathrm{d}t^{k}} \mathrm{z}_i = (R_{k-1}(\mathbf{z}(t), t))_i$ with $R_0(\mathbf{z},t) = \mathbf{v}(\mathbf{z}, t)$ and
\begin{equation*}
\begin{aligned}
     R_{k}(\mathbf{z}, t) \coloneqq \left(\sum_{j=0}^{d_z} \sum_{l=0}^{k-1} {k-1 \choose l} 
    \left(\frac{\partial}{\partial \mathrm{z}_j} R_{k-1-l}(\mathbf{z}, t)\right) (R_{l}(\mathbf{z}, t))_j\right) + \frac{\partial}{\partial t} R_{k-1}(\mathbf{z}, t).
\end{aligned}
\end{equation*}
Using the numerical solver values and these higher-order derivatives in combination with the Hermite interpolation will give us the following error estimate:

\begin{theorem}
    Assume we have a numerical ODE solver $\Psi$ with global error order $\mathcal{O}(\delta_t^q)$ and a time discretization $\bm{t} := \{t_i\}_{i=0}^{N_{\delta_t}-1}$ with $\sup_{i\in\{1, \ldots, N_{\delta_t}-1\}}|t_i - t_{i-1}| \leq \delta_t$. Furthermore, assume for all $k\leq p-1$ $R_k(\mathbf{z},t)$ is $L_k$-Lipschitz in $\mathbf{z}$ and assume
    the functions $(\mathbf{z}, t) \mapsto (R_{2p-1}(\mathbf{z}, t))_j$ are in $L^\infty(Z\times \R)$ for $j=1, \ldots, d_z$. Then, when using a Hermite polynomial of degree $2p-1$ in Equation \eqref{eq:full_numerical_solver_ODE}, the following error estimate applies,
    \begin{equation*}
        \norm{\mathcal{F}_t(\mathbf{z}) - \widehat{\mathcal{F}}_t(\mathbf{z};\bm{t})}_2 = \mathcal{O}\left(\delta_t^{\min(2p,q)}\right).
    \end{equation*}
\end{theorem}
\begin{proof}
    Take a time $t \in [t_i, t_{i+1}]$. Then $\mathcal{F}_t(\mathbf{z}):=\mathbf{z}(t)$ and $\widehat{\mathcal{F}}_t(\mathbf{z} ; \bm{t}) = \bm{s}_i(t)$. Define the numerical approximation $\widehat{\mathbf{z}}(\bm{t}):=\{\widehat{\mathbf{z}}(t_i)\}_{i=0}^{N_{\delta_t}-1}$ to Equation \eqref{eq:latent_ode}. Moreover, define $\tilde{\bm{s}}_i$ analogous to $\bm{s}_i$, but instead interpolating the ground truth $\mathbf{z}(t_i)$ and $\mathbf{z}(t_{i+1})$ values. Then for $t \in [t_i, t_{i+1}]$ we obtain
    \begin{equation}
        \begin{split}
            \norm{\mathbf{z}(t) - \bm{s}_i(t)}_2 & \leq \norm{\mathbf{z}(t) - \tilde{\bm{s}}_i(t)}_2 + \norm{\tilde{\bm{s}}_i(t) - \bm{s}_i(t)}_2 \\
            & \leq \frac{\delta_t^{2p}}{2^{2p}(2p)!} \left({\sum_{j=1}^{d_z} \norm{\frac{\mathrm{d}^{2p}}{\mathrm{d}t^{2p}} \mathrm{z}_j}_{L^\infty(t_i,t_{i+1})}^2} \right)^{\frac{1}{2}} + \norm{\tilde{\bm{s}}_i(t) - \bm{s}_i(t)}_2,
        \end{split}
        \label{eq:general_error_estimate_ode_solver_p_unequal_to_2}
    \end{equation} 
    where the final inequality follows from standard Hermite interpolation bounds \cite{stoer1980introduction, corless2013graduate}. The existence and $L^\infty(t_i,t_{i+1})$-boundedness of the $2p$-th time derivatives comes from  $\frac{\mathrm{d}^{2p}}{\mathrm{d}t^{2p}} \mathrm{z}_j = (R_{2p-1}(\mathbf{z}(t), t))_j$ and $(\mathbf{z}, t) \mapsto (R_{2p-1}(\mathbf{z}, t))_j$ being in $L^\infty(Z\times \R)$.
    
    By defining  $\delta_i:=(t_{i+1}-t_i)$, $w_{2p-1}(x):= \left[1, x, \ldots, x^{2p-1} \right]$, and $\tilde{t} = \frac{t -t_i}{\delta_i},$ the spline interpolation can be written as:
    \begin{equation*}
        s_{ij}(t) = w_{2p-1}(\tilde{t})A_{2p-1}\begin{bmatrix}
            \mathrm{z}_j(t_i) \\
            \mathrm{z}_j(t_{i+1}) \\
            \vdots \\
            \delta_i^k \frac{\mathrm{d}^k \mathrm{z}_j(t_i)}{\mathrm{d}t^k } \\
            \delta_i^k \frac{\mathrm{d}^k \mathrm{z}_j(t_{i+1})}{\mathrm{d}t^k }\\
            \vdots
        \end{bmatrix},
    \end{equation*}
    where $A_{2p-1}$ is the fixed interpolation matrix for Hermite interpolation on the interval $[0, 1]$. With this formulation for the spline interpolation, the interpolation error on each interval can be written as,
    {\allowdisplaybreaks
    \begin{align*}
            &\norm{\tilde{\bm{s}}_{i}(t) - \bm{s}_{i}(t)}_2\\ 
            & = \sqrt{\sum_{j=1}^{d_z} |\tilde{s}_{ij}(t) - s_{ij}(t)|^2} \\
            & = \sqrt{\sum_{j=1}^{d_z} |w_{2p-1}(\tilde{t})A_{2p-1} \begin{bmatrix}
            \mathrm{z}_j(t_i) - \widehat{\mathrm{z}}_j(t_i) \\
            \mathrm{z}_j(t_{i+1}) - \widehat{\mathrm{z}}_j(t_{i+1}) \\
            \vdots \\
            \delta_i^k \left(\frac{\mathrm{d}^k }{\mathrm{d}t^k }\mathrm{z}_j( t_i) - \frac{\mathrm{d}^k }{\mathrm{d}t^k }\widehat{\mathrm{z}}_j( t_i)  \right)  \\
            \delta_i^k \left(\frac{\mathrm{d}^k }{\mathrm{d}t^k }\mathrm{z}_j( t_{i+1}) - \frac{\mathrm{d}^k }{\mathrm{d}t^k }\widehat{\mathrm{z}}_j(t_{i+1})  \right)\\
            \vdots
        \end{bmatrix} |^2} \\
        & \leq \sqrt{\sum_{j=1}^{d_z} \norm{w_{2p-1}(\tilde{t})A_{2p-1}}_2^2 \norm{\begin{bmatrix}
            \mathrm{z}_j(t_i) - \widehat{\mathrm{z}}_j(t_i) \\
            \mathrm{z}_j(t_{i+1}) - \widehat{\mathrm{z}}_j(t_{i+1}) \\
            \vdots \\
            \delta_i^k \left(\frac{\mathrm{d}^k }{\mathrm{d}t^k }\mathrm{z}_j( t_i) - \frac{\mathrm{d}^k }{\mathrm{d}t^k }\widehat{\mathrm{z}}_j( t_i)  \right)  \\
            \delta_i^k \left(\frac{\mathrm{d}^k }{\mathrm{d}t^k }\mathrm{z}_j( t_{i+1}) - \frac{\mathrm{d}^k }{\mathrm{d}t^k }\widehat{\mathrm{z}}_j( t_{i+1})  \right)\\
            \vdots
        \end{bmatrix}}_2^2} \\
        & = \norm{w_{2p-1}(\tilde{t})A_{2p-1}}_2 \Bigg( \sum_{k=0}^{p-1} \norm{\delta_i^k \left(\frac{\mathrm{d}^k }{\mathrm{d}t^k }\mathbf{z}(t_i) - \frac{\mathrm{d}^k}{\mathrm{d}t^k}\widehat{\mathbf{z}}(t_i)\right) }_2^2 \\
          & \hfill + \norm{\delta_i^k \left(\frac{\mathrm{d}^k }{\mathrm{d}t^k }\mathbf{z}(t_{i+1}) - \frac{\mathrm{d}^k }{\mathrm{d}t^k }\mathbf{z}( t_{i+1})  \right)}_2^2 \Bigg)^{\frac{1}{2}} \\
        & = \norm{w_{2p-1}(\tilde{t})A_{2p-1}}_2 \sqrt{\sum_{k=0}^{p-1} \sum_{t \in \{t_i, t_{i+1}\}} \norm{\delta_i^k \left(R_{k-1}(\mathbf{z}(t), t) - R_{k-1}(\widehat{\mathbf{z}}(t), t)\right) }_2^2} \\
        & \leq \norm{w_{2p-1}(\tilde{t})A_{2p-1}}_2 \sqrt{\sum_{k=0}^{p-1} (L_k \delta_i^k)^2 (\norm{\mathbf{z}(t_i) - \widehat{\mathbf{z}}(t_i)}_2^2 + \norm{\mathbf{z}(t_{i+1}) - \widehat{\mathbf{z}}(t_{i+1})}_2^2)} \\
        & \leq \norm{w_{2p-1}(\tilde{t})A_{2p-1}}_2 \sqrt{\sum_{k=0}^{p-1} (L_k \delta_i^k)^2} \sqrt{(\norm{\mathbf{z}(t_i) - \widehat{\mathbf{z}}(t_i)}_2^2 + \norm{\mathbf{z}(t_{i+1}) - \widehat{\mathbf{z}}(t_{i+1})}_2^2)} \\
        & \leq \norm{w_{2p-1}(\tilde{t})A_{2p-1}}_2 \sqrt{\sum_{k=0}^{p-1} (L_k \delta_t^k)^2} \cdot \sqrt{2} C \delta_t^q,
    \end{align*}}
where the last inequality holds as the numerical solver is of the order $q$ and $\delta_i \leq \delta_t$. Combining this with inequality \eqref{eq:general_error_estimate_ode_solver_p_unequal_to_2} and with the boundedness of $w_{2p-1}(x)$ for $x \in [0, 1]$, we obtain that we are still of the order $\delta_t^{\min(2p,q)}$, hence proving our claim.
\end{proof}

\subsection{Koksma-Hlawka inequality for compact domains} \label{SM:koksma-hlawka_ineq}
\begin{theorem}[Koksma-Hlawka type error estimate for point cloud approximation of integrals \cite{pausinger2015koksma}] \label{thm:koksma_hlawka}
    Assume we have a set $\mathcal{D}$ of Borel sets $B \subset \Omega \subset \mathbb{R}^d$ and some measure $\mu$ on $\Omega$. Moreover, assume that $\mathcal{D}$ satisfies the following properties:
    \begin{itemize}
        \item $\mu(\partial B) = 0, \quad \forall B \in \mathcal
        D$.
        \item We have:
        \begin{equation*}
            \lim_{n \to \infty} \sup_{B \in \mathcal
            D} \left|\mu(B) - \frac{1}{n}\sum_{j=1}^n \mathbbm{1}_B(x_j)\right| = \lim_{n \to \infty} \operatorname{Disc}_{\mathcal{D}}(\{x_j\}_{j=1}^n) = 0
        \end{equation*}
        if and only if the infinite sequence $\{x_j\}_{j=1}^\infty$ satisfies
        \begin{equation*}
            \lim_{n \to \infty} \frac{1}{n} \sum_{j=1}^n f(x_j) = \int_\Omega f(\bm{x})\mu(dx), \quad \forall f \in C(\Omega).
        \end{equation*}
    \end{itemize}
    Define $\V_\infty(\mathcal{D}, \Omega)$ as the collection of all measurable functions $f \colon \Omega \rightarrow \R$ for which there exists a sequence $f_i = \sum_{j=1}^{n_i} \alpha_{ij} \mathbbm{1}_{A_{ij}}$ with $A_{ij} \in \mathcal{D}$ that converges uniformly to $f$. Moreover, define:
    \begin{equation*}
    \begin{aligned}
        V_\mathcal{D}(f) &\coloneqq \inf \Bigg( \liminf_i \left(\sum_{j=1}^{n_i} |\alpha_{ij}| \begin{cases}
            0, \quad \text{if } A_{ij} \in \{\emptyset, \Omega\} \\
            1, \quad \text{otherwise}
        \end{cases} \right)\\
        & \mid f_i= \sum_{j=1}^{n_i} \alpha_{ij} \mathbbm{1}_{A_{ij}}, A_{ij} \in \mathcal{D}, \lim_i \norm{f - f_i}_\infty = 0 \Bigg)
    \end{aligned}
    \end{equation*}
    For $f \in \V_\infty(\mathcal{D}, \Omega)$, we have:
    \begin{equation*}
        \left|\int_\Omega f(\bm{x}) \mu(dx) - \frac{1}{n}\sum_{i=1}^n f(x_j) \right| \leq V_\mathcal{D}(f) \operatorname{Disc}_\mathcal{D}(\{x_j\}_{j=1}^n)
    \end{equation*}
\end{theorem}

\section{Connection between RONOM's encoder and the CNO} \label{app:connection_encoder_and_CNO}

To handle inputs at different resolutions, the ReNO framework \cite{bartolucci2024representation} and the CNO \cite{Raonic2023} utilize frame sequences. 
\noindent\begin{definition}[Frame sequence]
    A countable sequence of elements $\phi_i \subset \mathcal{H}$ for some Hilbert space $\mathcal{H}$ and index set $I$ is called a frame sequence if there exists constants $A, B$ such that for all $u \in \mathcal{V}:=\overline{\mathSpan{(\phi_i : i \in I)}}$:
    \begin{equation*}
        A \norm{u}_{\mathcal{H}}^2 \leq \sum_{i \in I} |\braket{u}{\phi_i}|^2 \leq B \norm{u}_{\mathcal{H}}^2.
    \end{equation*}
\end{definition}
Given a frame sequence, we define the bounded linear synthesis operator $T \colon l^2(I) \to \mathcal{V}$ via $T(\{c_i\}) = \sum_{i\in I} c_i \phi_i$, and its adjoint, known as the analysis operator, via $T^*(u) = \{\braket{u}{\phi_i}\}_{i \in I}$. Given these definitions, there exists a projection of elements of $u \in \mathcal{H}$ onto $\mathcal{V}$ using the inverse of the \textit{frame operator} $S:=TT^*$:
\begin{equation}
    \mathcal{P}_\mathcal{V} u = T T^\dagger u = \sum_{i \in I} \braket{u}{S^{-1} \phi_i} \phi_i = \sum_{i \in I} \braket{u}{\phi_i} S^{-1} \phi_i.
    \label{eq:frame_proj}
\end{equation}
The CNO uses as frame (sequence) the orthonormal basis $\{\prod_{i=1}^d\sinc(2\omega \cdot - n_i) \mid n_i \in \mathbb{Z} \}$ of the space $\mathcal{B}_\omega(\mathbb{R}^d)$ of bandlimited functions. Reconstructions can be computed using values on a grid due to $\braket{f}{\prod_{i=1}^d\sinc(2\omega \cdot - n_i)}_{L_2(\Omega)} = f(..., \frac{n_i}{2\omega}, ...)$. One can also approximate the original $L_2$ inner products and use general meshes to remove the grid constraint. However, this implies calculating an infinite number of $L_2$ inner products.

To alleviate the latter issue, one can use a finite frame sequence $\{\phi_i\}_{i \in I}$ with $|I|<\infty$, which results in the following explicit projection formula.
\begin{theorem}[Projection onto a finite frame]
    For a finite frame $\{\phi_i\}_{i \in I}$ with $|I|<\infty$ the projection formula onto $\mathcal{V} = \overline{\mathSpan {(\phi_1,\ldots, \phi_n)}} = \mathSpan {(\phi_1, \ldots, \phi_n)}$ is,
    \begin{equation*}
       \mathcal{P}_\mathcal{V} u = \sum_{i=1}^{N_b} \left(\Phi^\dagger\braket{u}{\bm{\phi}}\right)_i  \phi_i,
    \end{equation*}
    where $\Phi_{ij}:=\braket{\phi_i}{\phi_j}$, $\left(\braket{u}{\bm{\phi}}\right)_i:= \braket{u}{\phi_i}$, and $\Phi^\dagger$ denotes the pseudo-inverse of $\Phi$.
\label{thm:proj_frame_formula}
\end{theorem}
\begin{proof}
    Assume $Sf = \phi_k$ for $f \in \mathcal{V}$. As $S$ is linear and $f = \sum_{j=1}^{N_b} \alpha_j \phi_j$, we get:
    \begin{equation}
        \begin{split}
            Sf & = \sum_{i=1}^{N_b} \braket{f}{\phi_i} \phi_i =  \sum_{i=1}^{N_b} \braket{\sum_{j=1}^{N_b} \alpha_j \phi_j}{\phi_i} \phi_i = \sum_{i=1}^{N_b} \left( \sum_{j=1}^{N_b}\braket{\phi_j}{\phi_i} \alpha_j \right)\phi_i \\
            & = \sum_{i=1}^{N_b} \left( \sum_{j=1}^{N_b} \Phi^T_{ij} \alpha_j \right)\phi_i  = \sum_{i=1}^{N_b} \left( \Phi^T\bm{\alpha} \right)_{i} \phi_i = \phi_k.
        \end{split}
        \label{eq:S_inv_determination_formula}
    \end{equation}
    Taking the inner product with $\phi_j$ gives us:
    \begin{equation*}
        \sum_{i=1}^{N_b} (\Phi^T \bm{\alpha})_i \braket{\phi_j}{\phi_i} = \braket{\phi_j}{\phi_k}.
    \end{equation*}
    Equivalently, this means that $\bm{\alpha}=[\alpha_1, \ldots, \alpha_n]$ satisfies the normal equations $\Phi \Phi^T \bm{\alpha} = \Phi e_k$ of the minimization of $\norm{\Phi^T \bm{\alpha} - e_k}^2 = \norm{\Phi \bm{\alpha} - e_k}^2$. This is a necessary condition. To see that it is also a sufficient condition, note that for any $\bm{\alpha}$ satisfying the normal equations, we have 
    \begin{equation*}
    \begin{split}
    \braket{\phi_j}{\sum_{i=1}^{N_b} (\Phi^T \bm{\alpha} - e_k)_i \phi_i} &=  \sum_{i=1}^{N_b} (\Phi^T \bm{\alpha} - e_k)_i \braket{\phi_j}{ \phi_i}\\
    &= \left(\sum_{i=1}^{N_b} (\Phi^T \bm{\alpha})_i\braket{\phi_j}{ \phi_i}\right) - \braket{\phi_j}{\phi_k} \\
    & = \braket{\phi_j}{\phi_k} - \braket{\phi_j}{\phi_k} = 0,
    \end{split}
    \end{equation*}
    where the second-to-last inequality follows from the normal equations. Consequently, $\braket{f}{\sum_{i=1}^{N_b} (\Phi^T \bm{\alpha} - e_k)_i \phi_i} = 0$ for all $f \in \mathSpan{(\phi_i \mid i=1, \ldots, {N_b})}$, which implies that $\sum_{i=1}^{N_b} (\Phi^T \bm{\alpha})_i \phi_i = \phi_k$ and showcasing that every solution to the normal equation solves Equation \eqref{eq:S_inv_determination_formula}. 
    
    Hence, one possible $\bm{\alpha}$ is given by the minimum-norm solution $\Phi^\dagger e_k$ of the minimization problem. Filling this into Equation \eqref{eq:frame_proj} yields
    \begin{equation*}
        \begin{split}
            \mathcal{P}_\mathcal{V} u & = \sum_{i = 1}^{N_b} \braket{u}{S^{-1} \phi_i} \phi_i = \sum_{i=1}^{N_b} \braket{u}{\sum_{j=1}^{N_b} (\Phi^\dagger e_i)_j \phi_j} \phi_i \\
            & = \sum_{i=1}^{N_b} \left(\sum_{j=1}^{N_b} (\Phi^\dagger)_{ij}\braket{u}{\phi_j}\right) \phi_i  = \sum_{i=1}^{N_b} \left(\Phi^\dagger\braket{u}{\bm{\phi}}\right)_i  \phi_i,
        \end{split}
    \end{equation*}
    which is the formula we set out to prove.
\end{proof}

The theorem above shows that the coefficients in the finite-frame expansion satisfy $\Phi \bm{\alpha} = \braket{u}{\bm{\phi}}$. Putting the gradient with respect to $\bm{\alpha}$ to zero in problem \eqref{eq:opt_prob_cond_proj}, the coefficients of our regularized projection $\P_\V^\lambda(u)$ are given by solutions of $(\Phi + \lambda L)\bm{\alpha} = \braket{u}{\bm{\phi}}_{L_2(\Omega)}$. Thus, our regularized projection can be interpreted as a finite frame reconstruction with an additional regularization parameter.

\section{Design choices RONOM} \label{SM:ronom_arch_details}
This section outlines the design choices we made for RONOM in the experiments. As shown in Equations \eqref{eq:encoder_formula} and its discretized form \eqref{eq:discretized_encoder}, the encoder consists of three main components: the operator $\mathcal{E}_\varphi$, the measurement operator $M$, and the basis functions $\{\phi_i\}_{i=1}^{N_b}$ used for the regularized projection in \eqref{eq:reg_proj} and its discrete version in \eqref{eq:discr_proj}. Additionally, a regularization term $\norm{\cdot}_\V^2$ must be chosen to complete the regularized projection.

Since the data lies on a grid, $M$ is defined as the evaluation functional at the training grid points. Given this grid structure, $\mathcal{E}_\varphi$ is chosen as a convolutional neural network, using one-dimensional convolutions for one-dimensional spatial domains and two-dimensional convolutions for two-dimensional domains. The basis functions $\{\phi_i\}_{i=1}^{N_b}$ are Gaussians centered around the grid points. As the 1D case is similar, assume we have a uniform 2D grid $\bm{x}_{ij}=[i \Delta x, j \Delta x]$ with $i, j \in \{0, \ldots, S\}$, $\bm{x}_{00}=[0,0]$, $\bm{x}_{SS}=[1,1]$, and $\Delta x = \frac{1}{S}$. In this case, the basis functions can be written as follows,
\begin{equation*}
    \phi_{ij}(\bm{x}) = \exp\left(\frac{\norm{\bm{x} - \bm{x}_{ij}}^2}{2\sigma_S^2}\right), \quad \sigma_S\coloneqq \frac{1}{S \sqrt{-2\ln(0.5)}}
\end{equation*}
Moreover, we add Gaussians outside the boundary. We add $N_p$ Gaussians along each edge, which introduces functions centered at positions $\bm{x}_{0j} - [k \Delta x, 0]$, $\bm{x}_{1j} + [k \Delta x, 0]$, $\bm{x}_{i0} - [0, k \Delta x]$, and $\bm{x}_{i1} + [0, k \Delta x]$ for $k=1, \ldots N_p$. 

The projection operator constructed using the chosen basis functions includes a regularization term $\norm{\cdot}_\V^2$. In the numerical experiments, a composite regularizer comprising three individual regularization terms is employed. For the first term, a Sobolev norm $\norm{\Delta u}_{L_2(\Omega)}^2$ is selected, which can be computed analytically for the function space $\V = \mathSpan(\phi_1, \phi_2, \ldots, \phi_{N_b})$. For the second term, an $L_2(\Omega)$-orthonormal basis $\{\tilde{\phi}_i\}_{i=1}^{N_b}$ of $\V$ is constructed via, e.g. via Gram-Schmidt. Define an $\mathcal{L}_1$ operator via linearity and $ \mathcal{L}_1 \phi_i \coloneqq \tilde{\phi_i}$. It leads to,
\begin{equation*}
    \norm{\L_1 u}_{L_2(\Omega)}^2 = \norm{\sum_{i=1}^{N_b} a_i \L_1 \phi_i}_{L_2(\Omega)}^2 = \sum_{i,j =1}^{N_b} a_i a_j \braket{\tilde{\phi}_i}{\tilde{\phi}_j}_{L_2(\Omega)} = \norm{\bm{a}}_2^2
\end{equation*}
The third term introduces an optional boundary condition operator $\mathcal{L}_2 u$, which maps to the residual error from a prescribed linear boundary condition evaluated at selected boundary points. The corresponding regularization term is given by $\norm{\L_2 u}_2^2$. 

Combining all terms leads to $\norm{u}_\V^2 \coloneqq \norm{\Delta u}_{L_2(\Omega)}^2 + \gamma_1 \norm{\L_1 u}_{L_2(\Omega)}^2 + \gamma_2 \norm{\L_2 u}_2 ^2$, where $\gamma_1$ and $\gamma_2$ are scalar weighting parameters.

The last component to define for the projection is the weighting vector $w$ shown in Equation~\eqref{eq:discr_proj}. Each entry of $w$ is determined by multiplying the base value of $1$ by $0.5$ for every boundary the corresponding grid point lies on: corner points receive a weight of $0.25$, edge points $0.5$, and interior points $1$.

The decoder $D_d$ in Equation \eqref{eq:decoder} maps the latent code to values on the grid $\bm{x}_{ij}$ ($i, j \in \{0, \ldots, S\}$) via a convolutional neural network. For the kernel in the optimal recovery problem in Equation \eqref{eq:decoder}, a Gaussian kernel with standard deviation $\sigma_S$ is used.

\section{Lipschitz continuity of neural networks and RONOM}\label{SM:lipschitz_cont_of_NNs_and_RONOM}
In the main manuscript, the neural networks under consideration are often assumed to be Lipschitz. While this may initially appear restrictive, many neural networks used in practice satisfy this property.

To see this, consider first a single-layer network $f_\theta \colon \R^d \to \R^n$ of the form
\begin{equation}\label{eq:one_layer_NN}
    f_\theta(\bm{x}) = \mathrm{A} \, \sigma(\mathrm{W}\bm{x} + b),
\end{equation}
where $\mathrm{A}$ and $\mathrm{W}$ are matrices, $b$ is a vector, and the activation function $\sigma$ is applied componentwise. Assume that $\sigma$ is $L_\sigma$-Lipschitz, which holds for most commonly used activation functions. Furthermore, abusing notation slightly, we denote by $\|\cdot\|$ generic vector and operator norms, even though they are applied to objects of different dimensions. With this convention, we obtain
\begin{equation*}
\begin{aligned}
    \|f_\theta(\bm{x}) - f_\theta(\bm{y})\|
    &= \big\|\mathrm{A}\big(\sigma(\mathrm{W}\bm{x} + b) - \sigma(\mathrm{W}\bm{y} + b)\big)\big\| \\
    &\le \|\mathrm{A}\| \, \big\|\sigma(\mathrm{W}\bm{x} + b) - \sigma(\mathrm{W}\bm{y} + b)\big\| \\
    &\le L_\sigma \|\mathrm{A}\| \, \|\mathrm{W}(\bm{x}-\bm{y})\| \\
    &\le L_\sigma \|\mathrm{A}\| \, \|\mathrm{W}\| \, \|\bm{x}-\bm{y}\|.
\end{aligned}
\end{equation*}
Thus, a single-layer neural network is Lipschitz whenever its activation function is.

For deeper architectures, the argument follows by induction. Suppose that
\begin{equation*}
    f_\theta(\bm{x}) = \mathrm{A} \, \sigma(\mathrm{W} g(\bm{x}) + b),
\end{equation*}
where $g$ is a network with one fewer layer and is $L_g$-Lipschitz by the induction hypothesis. Applying the single-layer estimate yields
\begin{equation*}
\begin{aligned}
    \|f_\theta(\bm{x}) - f_\theta(\bm{y})\|
    &\le L_\sigma \|\mathrm{A}\| \, \|\mathrm{W}\| \, \|g(\bm{x}) - g(\bm{y})\| \\
    &\le L_\sigma L_g \|\mathrm{A}\| \, \|\mathrm{W}\| \, \|\bm{x}-\bm{y}\|.
\end{aligned}
\end{equation*}
This completes the inductive step and demonstrates that standard feedforward neural networks are Lipschitz as soon as their activation functions are.

Here, we focused on typical pointwise activation functions and do not explicitly address mappings such as the softmax, which commonly appear in attention mechanisms. Since the softmax is also Lipschitz, a similar argument is expected to apply in that setting as well.

\subsection{Lipschitz continuity of RONOM}
The Lipschitz continuity of neural networks, as discussed previously, also implies that RONOM is Lipschitz. With a slight abuse of notation, we write $\norm{\cdot}$ for arbitrary vector norms. For the encoder defined in \eqref{eq:encoder_formula}, we obtain
\begin{align*}
    \norm{E(u_0) - E(\tilde{u}_0)} 
    &= \norm{\mathcal{E}_\varphi\!\left(M\P_\V^\lambda(u_0)\right) - \mathcal{E}_\varphi\!\left(M\P_\V^\lambda(\tilde{u}_0)\right)} \\
    &\le L_{\mathcal{E}_\varphi}\,\norm{M\!\left(\P_\V^\lambda(u_0 - \tilde{u}_0)\right)} \\
    &\le L_{\mathcal{E}_\varphi}\,\norm{M \P_\V^\lambda}_{L_2(\Omega)\to\R^{d_m}}
        \,\norm{u_0 - \tilde{u}_0}_{L_2(\Omega)} \\
    &\eqqcolon L_E \norm{u_0 - \tilde{u}_0}_{L_2(\Omega)},
\end{align*}
where $\norm{M \P_\V^\lambda}_{L_2(\Omega)\to\R^{d_m}}$ denotes the corresponding operator norm, which is finite since
\begin{equation*}
    \norm{M \P_\V^\lambda u}_{\R^{d_m}}
    \le \norm{M}_{\V\to\R^{d_m}} \norm{\P_\V^\lambda u}_\V
    \le C\,\norm{M}_{\V\to\R^{d_m}} \norm{\P_\V^\lambda u}_{L_2(\Omega)}
    < \infty.
\end{equation*}
Here, the second inequality follows from the equivalence of norms on finite-dimensional spaces, and the last inequality uses the bound
$\norm{\P_\V^\lambda}_{L_2(\Omega)} \le \norm{\P_\V}_{L_2(\Omega)} \le 1$.

In addition to the encoder, we assume that $D_d$ in \eqref{eq:decoder} is Lipschitz. By Theorem \ref{thm:D_lipschitzness}, this implies that the full decoder $D$ is Lipschitz as well. Moreover, by \cite[Theorem 2.8]{teschl2012ordinary}, we have
\[
    \norm{\mathcal{F}_t(\mathbf{z}) - \mathcal{F}_t(\mathbf{z}+\delta_z)}_2
    \le \norm{\delta_z}_2 e^{L_v t},
\]
which shows that, for fixed $t$ and an $L_v$-Lipschitz velocity field $\mathbf{v}$ in the neural ordinary differential equation \eqref{eq:latent_ode}, the flow map $\mathcal{F}_t$ is $L_{\mathcal{F}_t}$-Lipschitz with
\[
    L_{\mathcal{F}_t} \coloneqq e^{L_v t}.
\]

Combining these Lipschitz estimates yields
\begin{align*}
    \norm{\mathcal{K}^\dagger(u_0, t) - \mathcal{K}^\dagger(\tilde{u}_0, t)}
    &= \norm{\left(D \circ \mathcal{F}_t \circ E\right)(u_0)
      - \left(D \circ \mathcal{F}_t \circ E\right)(\tilde{u}_0)} \\
    &\le L_D \norm{\mathcal{F}_t(E(u_0)) - \mathcal{F}_t(E(\tilde{u}_0))} \\
    &\le L_D L_{\mathcal{F}_t} \norm{E(u_0) - E(\tilde{u}_0)} \\
    &\le L_D L_{\mathcal{F}_t} L_E \norm{u_0 - \tilde{u}_0}_{L_2(\Omega)}.
\end{align*}
Hence, the operator $\mathcal{K}^\dagger$ learned by RONOM is Lipschitz with respect to the $L_2(\Omega)$ norm.

\section{Boundedness of neural network derivatives}\label{SM:bounded_derivs_NNs}
We establish that derivatives of neural networks are bounded via a proof by induction on the number of layers. For a one-layer neural network $f_\theta(\bm{x})$ as in Equation \eqref{eq:one_layer_NN}, we have
\begin{align*}
    \frac{\partial^l}{\partial x_{i_1} \cdots \partial x_{i_l}} \left(f_\theta(\bm{x})\right)_i 
    &= \sum_{k} \mathrm{A}_{ik} \frac{\partial^l}{\partial x_{i_1} \cdots \partial x_{i_l}} \sigma \left(\sum_j \mathrm{W}_{kj}x_j + b_k \right) \\
    &= \sum_{k} \mathrm{A}_{ik} \sigma^{(l)} \left(\sum_j \mathrm{W}_{kj}x_j + b_k \right) \mathrm{W}_{k i_1} \cdots \mathrm{W}_{k i_l},
\end{align*}
which is clearly bounded if the $l$-th derivative of $\sigma$ is bounded. This holds for almost all commonly used activation functions.

Hence, for one-layer neural networks, the claim that the derivatives are bounded is satisfied. We now extend this result to $n$-layer neural networks by induction. Consider
\begin{equation*}
    \sigma\left( \sum_{j} \mathrm{W}_{kj} g(\bm{x})_j + b_k \right),
\end{equation*}
where $\sigma \colon \R \to \R$ is the activation function and $g(\bm{x})$ represents an $(n-1)$-layer neural network. Applying the multivariate Faà di Bruno formula \cite{hardy2006combinatorics} yields
\begin{align*}
    & \frac{\partial^l}{\partial x_{i_1} \cdots \partial x_{i_l}} \sigma\left( \sum_{j} \mathrm{W}_{kj} g(\bm{x})_j + b_k \right) \\
    &= \sum_{\pi} \sigma^{(|\pi|)}\left( \sum_{j} \mathrm{W}_{kj} g(\bm{x})_j + b_k \right) \prod_{B \in \pi} \frac{\partial^{|B|}}{\prod_{m \in B} \partial x_{m}} \left(\sum_{j} \mathrm{W}_{kj} g(\bm{x})_j + b_k \right) \\
    &= \sum_{\pi} \sigma^{(|\pi|)}\left( \sum_{j} \mathrm{W}_{kj} g(\bm{x})_j + b_k \right)  \prod_{B \in \pi}\sum_{j} \mathrm{W}_{kj} \frac{\partial^{|B|} g_j}{\prod_{m \in B} \partial x_{m}},
\end{align*}
where $\pi$ is a partition of the set $\{1, \ldots, d\}$ with $\bm{x}\in \R^d$ and $|\pi|$ denotes the number of sets in the partition. By the induction hypothesis, the derivatives of $g_j$ are bounded. Since $\sigma^{(|\pi|)}$ is assumed to be bounded, the entire expression is bounded. Finally,
\begin{equation*}
    \frac{\partial^l}{\partial x_{i_1} \cdots \partial x_{i_l}} \left(f_\theta(\bm{x})\right)_i 
    = \sum_{k} \mathrm{A}_{ik} \frac{\partial^l}{\partial x_{i_1} \cdots \partial x_{i_l}} \sigma \left(\sum_j \mathrm{W}_{kj} g(\bm{x})_j + b_k \right),
\end{equation*}
which is bounded by our previous observation.

In summary, if the derivatives of the activation function are bounded up to the desired order, the higher-order derivatives of a neural network are bounded.  

\bibliographystyle{siamplain}
\bibliography{references}